\documentclass[11pt]{article}
 
\usepackage[final]{acl}

\usepackage{times}
\usepackage{latexsym}
\usepackage{amsmath}
\usepackage{amssymb}
\usepackage{float}
\usepackage{tikz}
\usetikzlibrary{positioning, arrows.meta, calc, shadows, backgrounds}
\definecolor{cotgrey}{RGB}{95,95,100}
\definecolor{loopblue}{RGB}{31,90,180}
\definecolor{tspteal}{RGB}{0,132,128}
\definecolor{redamber}{RGB}{242,181,44}
\definecolor{depgrey}{RGB}{178,178,183}
\usepackage{xcolor}
\usepackage[most]{tcolorbox}
\usepackage[T1]{fontenc}

\usepackage[utf8]{inputenc}

\usepackage{microtype}

\usepackage{inconsolata}

\usepackage{graphicx}
\usepackage{fontawesome5}

\usepackage{booktabs}

\usepackage{multirow}

\usepackage{comment}
\usepackage{todonotes}

\usepackage[ruled,vlined,linesnumbered]{algorithm2e}
\makeatletter
\newcommand{\removelatexerror}{\let\@latex@error\@gobble}
\makeatother

\usepackage{enumitem}

\usepackage{tabularx}

\usepackage{fancyvrb}

\definecolor{rqbg}{RGB}{246,246,246}
\definecolor{rqborder}{RGB}{125,125,125}
\definecolor{findingbg}{RGB}{235,248,239}
\definecolor{findingborder}{RGB}{64,145,86}
\newtcolorbox{questionboxstyle}{
  colback=rqbg, colframe=rqborder, boxrule=0.55pt,
  arc=2pt, left=3pt, right=3pt, top=2pt, bottom=2pt,
  before skip=4pt, after skip=4pt}
\newtcolorbox{findingboxstyle}{
  colback=findingbg, colframe=findingborder, boxrule=0.55pt,
  arc=2pt, left=3pt, right=3pt, top=2pt, bottom=2pt,
  before skip=4pt, after skip=4pt}
\newcommand{\ResultBox}[3]{\begin{findingboxstyle}\small\textbf{#1.} #2\\\textbf{Finding.} #3\end{findingboxstyle}}

\newcommand{\BLUE}[1]{#1}

\title{Understanding Failures in LLM Reasoning\\ by Learning Structured Representations of Chain-of-Thought}

\author{
  Tommaso Felice Banfi\thanks{Equal contribution}
  \quad Mubashara Akhtar$^*$ \quad Jingwei Ni \quad Mrinmaya Sachan \\
  ETH Zurich \\
  \texttt{\{tbanfi,makhta,jingni,msachan\}@ethz.ch}
}

\begin{document}
\maketitle

\begin{abstract}
Chain-of-Thought (CoT) is central to improving reasoning in large language models, yet existing evaluation methods focus on step-level signals and ignore the underlying structure of reasoning traces. This raises a key question: what aspects of reasoning quality (i.e., correctness, efficiency, and failure patterns) are encoded in the structure of a CoT trace?
We introduce \textbf{CoT2Graph}, a framework that represents CoT traces as directed dependency graphs, making relationships between reasoning steps explicit and capturing patterns such as backtracking, branching, and subtask merging. 
We model these graphs with graph neural networks and use attribution-based explanations to identify structural patterns associated with reasoning quality.
CoT2Graph reveals distinct patterns of successful and failed reasoning: correct CoTs place greater importance on early planning, whereas incorrect CoTs are associated with overloaded synthesis steps and dependency violations. We further show that graph-based signals better predict whether a CoT leads to a correct answer, outperforming previous tree-based structural methods by up to 2.5\% and simple length-based baselines by 7.8\%, while also improving Best-of-$N$ selection.
Overall, our results demonstrate that reasoning structure encodes rich signals about reasoning quality, motivating a shift from sequence-level evaluation to structural analysis.
\href{https://github.com/BanfiTommasoFelice/LLM-StructReasoning-EMNLP26}{\faGithub{\scriptsize{ BanfiTommasoFelice/LLM-StructReasoning-EMNLP26}}}

\end{abstract}


\section{Introduction}
\label{sec:introduction}

Models trained with reinforcement learning on reasoning tasks 
produce long Chain-of-Thought (LCoT) traces that involve backtracking over earlier conclusions, branching across concurrent sub-problems, and combining partial results from multiple reasoning threads~\cite{wu2026when}.
However, existing evaluation approaches treat these traces either as black boxes, measuring only the final-answer correctness, or as linear sequences, scoring each step individually. 
\begin{figure}[!t]
	\centering
    \text{CoT2Graph: CoT $\to$ Reasoning Structure}
	\includegraphics[width=\linewidth]{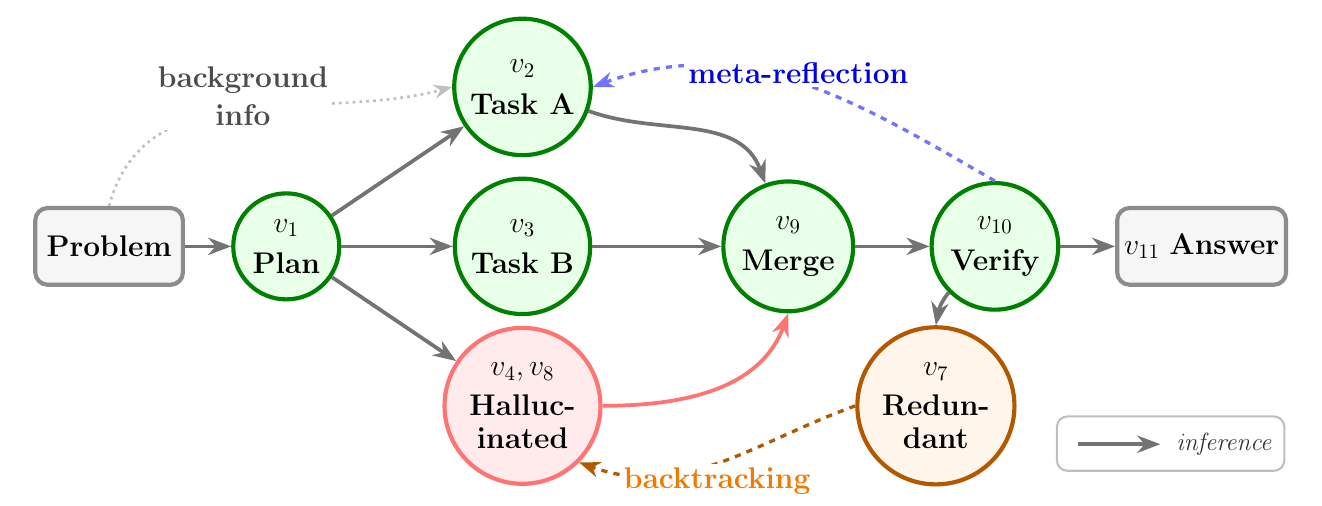}
	\caption{A CoT chain represented as an interpretable directed graph capturing the reasoning process.} 
	\label{fig:intro_graph}
\end{figure}
This raises a central question: what aspects of \emph{reasoning quality} such as correctness, efficiency, and failure patterns are encoded in the structure of a CoT trace?
Step-level evaluation frameworks such as ROSCOE~\cite{golovneva2023roscoe} and ReCEval~\cite{prasad-etal-2023-receval} detect local errors, but evaluate reasoning steps largely in isolation and therefore do not capture global reasoning structure.
Existing structured representations based on sequences or trees~\cite{dalvi-etal-2021-explaining,jiang-etal-2025-makes} provide a more global view, but cannot represent multi-source dependencies, the merging of reasoning threads, or the revisitation and reuse of intermediate results. It remains unclear whether and how CoT structure encodes reasoning quality.

\begin{figure*}[!t]
	\centering
	\includegraphics[width=1.0\linewidth]{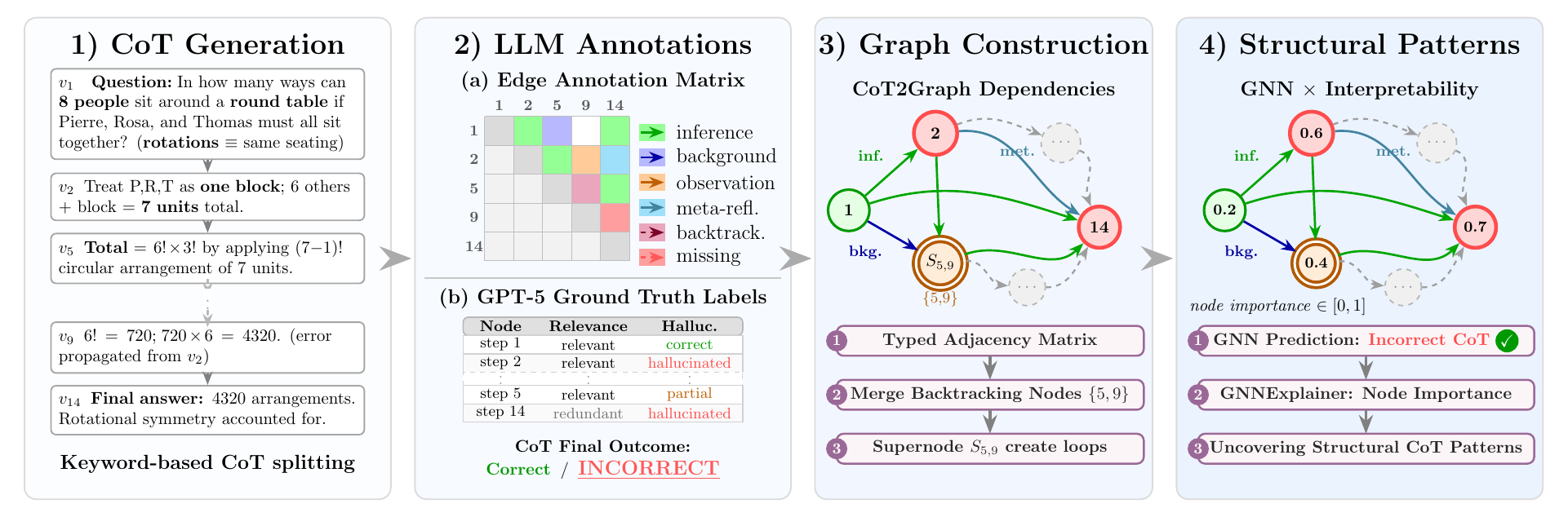}
	\caption{%
		CoT2Graph pipeline for graph construction and reasoning structural analysis (GPQA ID~32; Figure~\ref{fig:cot_graph}).
	}
	\label{fig:CoT2Graph}
\end{figure*}

To address this question, we introduce \textbf{CoT2Graph}, a framework that represents CoT traces as directed dependency graphs \BLUE{(Figure~\ref{fig:intro_graph})} and models them using graph neural networks (GNNs). Nodes are reasoning steps, and edges encode relations such as \textit{inference}, \textit{repetition}, \textit{observation}, and \textit{meta-reflection}.
The GNN predicts final-answer correctness from these typed dependencies, while GNNExplainer identifies the reasoning steps most decisive for each prediction. 
We provide new insights into reasoning behavior, 
%
identifying structural patterns of successful and failed reasoning traces. Correct CoTs place greater importance on early planning and task decomposition while incorrect CoTs, in contrast, are associated with overloaded synthesis steps that combine multiple reasoning threads and with violations of inferred dependencies, such as proceeding before prerequisite steps are generated. We also find that longer reasoning is not always more efficient: while longer traces contain more verification and exploration, they  are often more inefficient. These findings suggest that reasoning quality depends not only on the correctness of steps, but also on how intermediate reasoning is organized and composed in the CoT.

\textbf{Contributions:}
\begin{itemize}[leftmargin=*, nosep]
    \item We introduce CoT2Graph, which maps CoT traces to typed dependency graphs (\S\ref{sec:methodology}).
    \item Graph structure predicts correctness better than tree baselines, with gains up to 2.5\% (\S\ref{subsec:outcome}).
    \item Attributions reveal failure patterns: correct traces emphasize early planning, while incorrect traces rely on overloaded synthesis nodes (\S\ref{subsec:interpretation}).
    \item We decompose redundancy into verification--exploration patterns and path inefficiency, finding 49.0\% average path-efficiency loss (\S\ref{subsec:efficiency}).
\end{itemize}

\section{Related Work}
\label{sec:related}

\paragraph{Structured Representations of Reasoning.}
Recent reinforcement learning and test-time scaling approaches have enabled LLMs to generate longer CoT traces, leading to substantial improvements in reasoning performance~\cite{Guo_2025,fu2025deepthinkconfidence,chen2025research,ni2026reprobeefficienttesttimescaling}. However, most prior work focuses on improving performance rather than understanding the underlying reasoning structure.
Reasoning representations have been studied with entailment trees~\cite{dalvi-etal-2021-explaining}, which organize multi-hop inference into hierarchical steps.
Tree of Thoughts~\cite{10.5555/3666122.3666639,banfi2026llmsgametheoryentropyguided} and Graph of Thoughts~\cite{10.1609/aaai.v38i16.29720} impose structured reasoning at inference time via branching search.
\citet{jiang-etal-2025-makes} take the latter approach with LCoT2Tree, converting generated CoT traces to hierarchical trees via LLM-annotated sketches.
We extend this by using directed graphs that encode cycles and multi-source synthesis (inexpressible in a tree), while requiring a 7B model rather than a frontier LLM.

\paragraph{Step-Level Evaluation and Error Taxonomy.}
Step-level semantic frameworks evaluate CoT quality beyond final accuracy: ROSCOE~\cite{golovneva2023roscoe} covers alignment, coherence, and grammaticality; ReCEval~\cite{prasad-etal-2023-receval} assesses correctness and informativeness via NLI; and a recent survey~\cite{lee-hockenmaier-2025-evaluating} proposes a taxonomy of factuality, validity, coherence, and utility.
Error analysis has been applied to mathematical~\cite{li-etal-2024-evaluating-mathematical}, legal~\cite{mishra-etal-2025-investigating}, and multimodal CoT~\cite{jiang2025mmecot}, and verification methods correct intermediate errors via chain reversal~\cite{xue2023rcotdetectingrectifyingfactual}, external knowledge editing~\cite{zhao-etal-2023-verify}, and step-level verification~\cite{ijcai2025p925,jacovi2024chain}.
On efficiency, longer traces do not monotonically improve accuracy and can introduce redundancy~\cite{wu2026when,NEURIPS2024_01025a4e}.
Our work adds two structural error categories to the semantic taxonomy, shows that step redundancy is predictable from CoT structure, and provides an efficiency decomposition that separates productive exploration-verification patterns from inefficiency.


\section{CoT2Graph Framework}
\label{sec:methodology}

\textbf{CoT2Graph} converts Chain-of-Thought (CoT) traces into structured dependency graphs and models them using graph neural networks (GNNs) to analyze reasoning quality. The pipeline \BLUE{(Figure~\ref{fig:CoT2Graph})} consists of four stages: \textit{(i)} generating CoT traces and segmenting them into atomic reasoning steps, \textit{(ii)} annotating dependencies between nodes, and \textit{(iii)} constructing a graph representation for (iv) interpretability analysis.
This representation captures non-linear reasoning patterns such as branching, merging, and revisitation, while the GNN models \BLUE{support the analysis of} reasoning quality, including correctness, efficiency, and failure modes.

\subsection{Generation of CoT Traces}
We generate long CoTs using three open-weight reasoning models: DeepSeek-R1-Distill-Qwen-7B (DS-Qw7B)~\cite{Guo_2025}, DeepSeek-R1-Distill-Qwen-14B (DS-Qw14B), and MiMo-7B-RL~\cite{xiaomi2025mimo}. Traces are generated on the benchmarks GPQA and MMLU-Pro. 
To ensure comparability across models, we use a consistent setup with low temperature (T = $0.1$) and a maximum generation of $16$K tokens. This configuration produces stable yet sufficiently diverse CoT traces for downstream analysis.\footnote{Further details are provided in \S\ref{subsec:setup}.}

\subsection{Graph Representation}

We represent each CoT trace as a directed graph $G = (V, E, \phi, \psi)$, where nodes $V = \{v_1, \dots, v_n\}$ represent atomic reasoning steps and edges $E \subseteq V \times V$ capture dependencies between them.
Each edge is assigned a reasoning relation (i.e., \textit{inference}, \textit{background knowledge}, \textit{observation}, \textit{meta-reflection}, or \textit{repetition}) and a strength score indicating the importance of the dependency. 

For each node, the feature map stores its generation position and local topology, \BLUE{following the node features of LCoT2Tree~\cite{jiang-etal-2025-makes}}.

%
%
Let $\mathcal{T}$ be the edge-type vocabulary. \BLUE{Each edge $(v_i,v_j)$ is encoded by $\psi$ as its type $\tau_{ij} \in \mathcal{T}$ in one-hot form, concatenated with its normalized strength $s_{ij}$.}
Together, $\phi$ and $\psi$ transform a linear CoT into a typed, weighted dependency graph that exposes multi-step dependencies and non-linear reasoning patterns. To isolate structural signals, the predictor excludes CoT textual information and relies only on the CoT structure; step semantics are included as an ablation in Appx.~\ref{app:ablation_review}.
This representation preserves the order of reasoning steps, allowing us to study whether the organization of reasoning provides predictive signal about reasoning quality.


\subsection{Node Segmentation}

Each CoT trace is segmented into atomic reasoning steps $V = \{v_1, \dots, v_n\}$ using a rule-based parser that identifies structural transition markers (e.g., \textit{Therefore}, \textit{However}, and \textit{Finally}) via regex matching. Fragments shorter than 100 characters are merged to produce meaningful reasoning units.
The resulting sequence preserves the original generation order while producing coherent reasoning units for downstream analysis.
Appx.~\ref{app:segmentation_review} compares parser boundaries with GPT-5 Mini on GPQA and reports sensitivity across domains; DS-Qw7B reaches 91.5\% agreement, and 86.7\% of traces are within a 10\% boundary-error tolerance.

\subsection{Edge Annotation}

We infer dependencies between reasoning steps using an instruction-tuned LLM (Llama-3.1-8B-Instruct). For each ordered pair of steps $(v_i, v_j)$ with $i < j$, the model predicts \textit{(i)} a dependency type and \textit{(ii)} a strength/importance score $\in [0, 10]$, producing a typed directed edge.

The dependency type is selected from a fixed set of reasoning transitions: \textit{inference} (\BLUE{$v_j$ follows directly from $v_i$}), \textit{background} (\BLUE{$v_i$ supplies knowledge or assumptions used in $v_j$}), \textit{observation} (\BLUE{$v_j$ observes or comments on $v_i$}), \textit{meta-reflection} (\BLUE{$v_j$ plans or checks the reasoning rather than advancing it}), \textit{repetition} (\BLUE{$v_i$ and $v_j$ express the same content}), and \textit{none} (no meaningful dependency). Appx.~\ref{app:edge_type_definitions} gives the full definition and strength scale for each edge type, and relates the taxonomy to entailment, retrieval, and search-based reasoning literature.
Restricting to pairs with $i < j$ ensures that edges follow the original generation order, i.e., dependencies are only assigned from earlier to later steps. This encodes a causal prior consistent with how CoT traces are produced, where each step conditions on previously generated content.
The resulting edges capture how reasoning steps interact, including logical dependencies, incorporation of prior knowledge, and revisitation of earlier conclusions.

\subsection{Graph Construction and Merging}

The final reasoning graph combines the segmented nodes and annotated edges. 
\BLUE{A trace often states the same intermediate result more than once, so the same content would appear as several nodes and inflate the graph. We therefore compress repeated steps before building the graph.}
A \textit{repetition} edge defines semantic equivalence: its endpoints express the same content or a close rephrasing. Union--Find merges connected repetition components into equivalence classes.
%
%
Node features are recomputed on the compressed graph \BLUE{$G^{\star}$}, and edge features come from the retained compressed edges. 
\BLUE{When several edges collapse onto the same pair we keep the strongest one; equal-score ties use the fixed ordering background $\prec$ observation $\prec$ meta-reflection $\prec$ inference.} \BLUE{Merging} collapses repeated reasoning steps while preserving external dependencies; because equivalence classes can redirect formerly forward edges across merged components, the resulting graph can contain cycles that capture revisitation and verification behavior. 
Appx.~\ref{app:edge_merge_review} quantifies merging and tie resolution: 10.8\% of nodes participate in repetition components, and 75.1\% of equal-strength conflicts are same-type redirects, \BLUE{so the ordering is trivial in most cases}.

\section{Graph Modeling and Interpretation}

\BLUE{We obtain the labels that supervise training (\S\ref{subsec:labels}), train a GNN that predicts correctness from a CoT graph (\S\ref{subsec:gnn}), and use GNNExplainer for interpretability, attributing predictions back to nodes to identify which structures matter (\S\ref{subsec:interpretability}).}

\subsection{Step-Level Supervision Labels}
\label{subsec:labels}
We use GPT-5 Mini~\cite{singh2026openaigpt5card} as an LLM judge to obtain step- and trace-level labels along three dimensions:
\emph{(i) Relevance} categorizes whether a step advances the reasoning process (\textit{relevant}, \textit{redundant}, or \textit{irrelevant}).
\emph{(ii) Hallucination} evaluates the internal validity of each step in isolation (\textit{correct}, \textit{partial}, or \textit{hallucinated}).
\emph{(iii) Final-answer correctness} is a correct/incorrect label.
\BLUE{These labels serve as} (i)~supervision signals for the GNN models (\S\ref{subsec:outcome}), (ii)~pruning labels for the efficiency analysis in \S\ref{subsec:efficiency}, and (iii)~evaluation labels for Best-of-$N$ selection in \S\ref{subsec:bon}.
Thus, GPT-5 Mini defines target labels and Llama defines graph edges, but the correctness GNN receives only structural features and typed dependencies, not node text or judge labels. The experiment asks whether structure alone is sufficient to predict LLM-failure phenomena: a positive result indicates that the signal is encoded in reasoning topology, while a negative result means the signal may depend on node semantics, model or data scale, or may not be expressible from the graph alone.
Appx.~\ref{app:annotation_validation_review} reports judge--ensemble agreement of \BLUE{85.3\%} and Cohen \BLUE{$\kappa=0.82$} \BLUE{over 95 edges}; \BLUE{Appx.~\ref{app:prompts} lists all prompts.}

\subsection{GNN Modeling}
\label{subsec:gnn}


\BLUE{The model is given only structure: node features and typed edges, with no step text. This isolates CoT structure from the semantic content of the steps, so whatever the model predicts must come from how the reasoning is organized.} Given a compressed graph $G^{\star}$, a GNN~\cite{xu2018how} maps node features and typed edges to prediction scores. Let $h_i^{(0)}=\phi(v_i)$ and let $\widetilde{\mathcal{N}}^{-}(i)=\{i\}\cup\{j:(v_j,v_i)\in E\}$. The graph-level GATv2 encoder~\cite{brody2022how} updates node states as
\begin{equation}
\begin{aligned}
h_i^{(\ell+1)} &= \sigma\!\left(\sum_{j\in\widetilde{\mathcal{N}}^{-}(i)}
\alpha_{ji}^{(\ell)} W^{(\ell)}h_j^{(\ell)}\right),\\
\alpha_{ji}^{(\ell)} &= \operatorname{softmax}_{j\in\widetilde{\mathcal{N}}^{-}(i)}
 a_{\theta}^{(\ell)}(h_j^{(\ell)},h_i^{(\ell)},\bar{\psi}_{ji}).
\end{aligned}
\label{eq:gat_message_passing}
\end{equation}
Here $a_{\theta}^{(\ell)}$ is the GATv2 attention score \BLUE{(we use GATv2 as a strong standard choice for graph classification)}
, $\alpha_{ji}^{(\ell)}$ is normalized over the incoming neighborhood, and $\bar{\psi}_{ji}$ is the edge feature \BLUE{described above}, with a self-loop feature when $j=i$. After $K=2$ layers, mean pooling and an MLP give the outcome distribution
\begin{equation}
\begin{aligned}
z_{G^{\star}} &= \frac{1}{|V|}\sum_{i\in V}h_i^{(K)},\\
p_{G^{\star}} &= \operatorname{softmax}(\operatorname{MLP}(z_{G^{\star}})) \;=\; f_\theta(G^{\star}).
\end{aligned}
\label{eq:graph_prediction}
\end{equation}
For node-level analysis, a two-layer GINE encoder~\cite{hu2019strategies} uses the same graph representation with two heads, producing $p_i^{\mathrm{rel}}$ for relevance and $p_i^{\mathrm{hall}}$ for hallucination at each step $v_i$. These judge labels are supervision targets, not inputs to the main correctness classifier. The predictor is permutation-equivariant before pooling and permutation-invariant after pooling; Appx.~\ref{app:ablation_review} compares it with graph-statistic, MLP, sequence, and feature/edge ablations.


\subsection{\BLUE{Which Steps Carry the Outcome Signal}}
\label{subsec:interpretability}


\BLUE{To identify recurring patterns in CoT structure, we need to know which parts of a graph a prediction relies on. GNNExplainer~\cite{10.5555/3454287.3455116} answers this by learning a soft mask over nodes and edges that preserves the prediction while keeping the mask small.} We apply it to the trained predictor $f_\theta$, keeping its weights fixed.
  It optimizes a continuous node mask $m\in[0,1]^{|V|}$ that preserves the original prediction while keeping the explanation sparse and interpretable:
\begin{equation}
\begin{aligned}
m^{\star}=\arg\min_m \;& \mathcal{L}\!\left(f_\theta(G^{\star}\odot m),\hat{y}_{G^{\star}}\right)\\
&+\lambda_1\|m\|_1+\lambda_2 H(m).
\end{aligned}
\label{eq:gnn_explainer}
\end{equation}
Here $\hat{y}_{G^{\star}}=\arg\max_c p_{G^{\star}}(c)$, $\mathcal{L}$ is cross-entropy with target $\hat{y}_{G^{\star}}$, and $H(m)$ is mask entropy. This regularized objective approximates maximizing mutual information $\mathrm{MI}(Y,G^{\star}\odot m)$, where $Y$ is the GNN prediction random variable. We use node masks only, so $m_i^{\star}$ scores the importance of step $v_i$\BLUE{; since these scores are not on a common scale across graphs, we normalize each by its mean mask:} 
\begin{equation}
\tilde{m}_i=\frac{m_i^{\star}}{|V|^{-1}\sum_{u\in V}m_u^{\star}}.
\label{eq:normalized_node_mask}
\end{equation}

Thus $\tilde{m}_i>1$ denotes above-average importance within the same trace. These attributions explain the GNN predictor, not the LLM; we aggregate them by position, degree balance, and predicted outcome. Implementation details are in Appx.~\ref{app:implementation}.

\section{Results and Discussion}
\label{sec:experiments}

We evaluate whether CoT structure provides information about reasoning quality. The central question is not only whether a trace is correct, but which structural properties of the CoT are related to correctness, redundancy, hallucination, and useful verification.
Specifically, our experiments address five questions:
whether graph structure predicts final-answer correctness better than LR and LCoT2Tree (\S\ref{subsec:outcome}); which step-level phenomena are structurally visible (\S\ref{subsec:node_analysis}); what GNNExplainer reveals about successful and failed reasoning (\S\ref{subsec:interpretation}); how much long-CoT redundancy is useful verification rather than path inefficiency (\S\ref{subsec:efficiency}); and whether structural signals improve Best-of-$N$ selection (\S\ref{subsec:bon}).

\subsection{Experimental Setup}
\label{subsec:setup}

\paragraph{Evaluation Datasets and LLMs.}
Primary experiments use GPQA~\cite{rein2024gpqa}, a graduate-level
science benchmark requiring multi-step reasoning (physics, chemistry, and biology)
and 600 questions from MMLU-Pro~\cite{DBLP:conf/nips/WangMZNCGRAHJLK24}.
We evaluate three open-weight reasoning models spanning controlled variation in scale and post-training strategy:
DS-Qw7B and DS-Qw14B~\cite{Guo_2025},
trained via knowledge distillation from DeepSeek-R1, and
MiMo-7B-RL~\cite{xiaomi2025mimo}, trained with direct reinforcement learning on verifiable rewards. Appx.~\ref{app:cross_model} summarizes post-training differences among DS-Qw7B, DS-Qw14B, and MiMo-7B-RL, and reports the transfer setting used to test whether graph signals generalize across generators.

\subsection{\BLUE{Structure Encodes Answer Correctness}}
\label{subsec:outcome}

\ResultBox{Question}{\BLUE{Does the structure of a CoT dependency graph carry predictive signal for final-answer correctness?}}{
CoT2Graph consistently improves correctness prediction, outperforming LCoT2Tree by up to $2.5\%$ and LR by up to $7.8\%$.
}

\BLUE{We frame the task as binary graph-level classification.}
Across all evaluation settings, CoT2Graph consistently outperforms LCoT2Tree, achieving gains of up to $+2.5\%$ (see Table~\ref{tab:gnn_results}).
The advantage over LCoT2Tree is visible across all model families, demonstrating that
the richer representational capacity of graphs (which can encode cycles, 
cross-step dependencies, and non-hierarchical chains) translates into predictive gains.
We observe the largest margin over LCoT2Tree for DS-Qw14B ($+2.5\%$), suggesting that a richer graph representation better captures its more complex reasoning.

A logistic regression on node count, edge count, density, and degree statistics performs below the length-only baseline. This suggests that the gain is not due only to graph size or density. Appx.~\ref{app:ablation_review} reports graph-statistic baselines, pooled and sequence encoders, and edge-type/node-feature ablations.

\begin{table}[!t]
	\centering
	\small
	\setlength{\tabcolsep}{3.8pt}
	\renewcommand{\arraystretch}{1.15}
	\begin{tabular}{lccc}
		\toprule
		\textbf{Model}
		 & LR
		 & LCoT2Tree
		 & \textbf{CoT2Graph}               \\
		 & 
		 & ~\cite{jiang-etal-2025-makes}
		 & \textit{(ours)}                   \\
		\midrule
		DS-Qw7B
		 & 65.9
		 & 68.2 \scriptsize{(+2.3)}
		 & \textbf{70.5} \scriptsize{(+4.6)} \\
		DS-Qw14B
		 & 57.6
		 & 62.9 \scriptsize{(+5.3)}
		 & \textbf{65.4} \scriptsize{(+7.8)} \\
		MiMo-7B-RL
		 & 61.4
		 & 63.0 \scriptsize{(+1.6)}
		 & \textbf{64.8} \scriptsize{(+3.4)} \\
		\bottomrule
	\end{tabular}
    \caption{%
		Final-answer correctness prediction accuracy (\%) on GPQA. Parenthetical values indicate improvement over the Logistic Regression (LR) baseline. Our CoT2Graph consistently outperforms LCoT2Tree.
	}
	\label{tab:gnn_results}
\end{table}

\subsection{Structure Reveals Redundancy but Not Hallucination}
\label{subsec:node_analysis}
\ResultBox{Question}{Which CoT step-level properties are structural?}{
Graph structure predicts whether a step is redundant, but carries little signal about hallucination, suggesting that intra-step and semantic failures require complementary signals
(see \S\ref{subsec:bon}).}

Beyond final-answer prediction, we investigate whether graph topology encodes information
about the quality of individual reasoning steps, and which structural properties the
GNN uses to distinguish correct from incorrect CoT traces. 

\paragraph{Relevance is an Inter-Step Structural Property.}
Our node-level GNN trained to classify 
\textit{relevant}, \textit{redundant}, or \textit{irrelevant} steps achieves 65.2\% class accuracy on relevant nodes and 66.5\% on redundant nodes (see Table~\ref{tab:relevance_cm}), indicating that graph structure encodes a signal for detecting redundancy. Irrelevant nodes account for less than 1\% of the dataset and are grouped with redundant nodes. 

\paragraph{Hallucination is an Intra-Step Phenomenon.}
The graphs that predict \BLUE{final-answer correctness} and redundancy carry almost no signal about hallucinated\footnote{Appx.~\ref{app:prompts} defines a hallucination as an error within a step, such as an incorrect computation or unsupported claim, rather than an error in its dependencies.} steps.
Class accuracy falls to $7.0\%$ on partially hallucinated nodes and $7.9\%$ on \BLUE{hallucinated ones} (Appx.~\ref{app:node_hallucination_metrics}), against $65.2\%$ and $66.5\%$ for relevance under the same architecture.
We read this gap as a finding rather than a shortcoming: it marks where dependency structure stops being informative. Relevance is a property of how a step relates to its neighbors and is therefore visible in the topology. 
In contrast, hallucination is primarily an intra-step semantic property and is therefore poorly captured by our dependency representation.
The separation is also exploitable. Our CoT2Graph correlates weakly with entropy-based methods ($\rho=-0.16$), and the ensemble gains $1.6$ points in Best-of-$N$ (\S\ref{subsec:bon}): structural and semantic verifiers should be ensembled.

\subsection{\BLUE{Indicative Structural Patterns within CoT}}
\label{subsec:interpretation}
\ResultBox{Question}{CoT structure encodes predictive signal, but which structural patterns separate incorrect/correct CoTs?}{Correct CoTs rely on early planning; incorrect CoTs contain synthesis sinks merging many subprocesses.}

\paragraph{GNNExplainer (\S\ref{subsec:interpretability}) reveals two complementary structural patterns.} First, correct CoTs place greater predictive importance on early planning and task-decomposition steps. Second, incorrect CoTs place greater importance on intermediate, high-in-degree synthesis nodes that combine multiple reasoning threads.

\paragraph{Early Planning Steps Predict Correctness.}

\begin{figure}[!t]
	\centering
	\includegraphics[width=\linewidth]{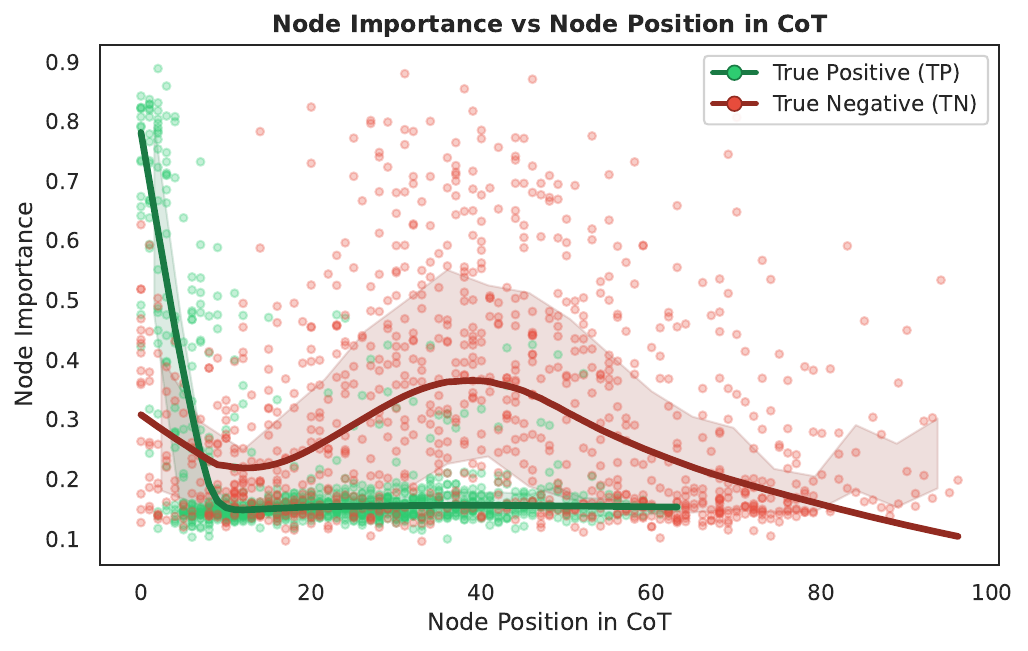}
	\caption{%
		GNNExplainer node importance across CoT node positions (GPQA, DS-Qw7B).
		Early nodes are more important in correct CoTs, whereas intermediate nodes are more important in incorrect CoTs.
	}
	\label{fig:nodeWeightvsPosition}
\end{figure}

In true positive predictions, node importance concentrates on the first 20--25\% of the steps (Figure~\ref{fig:nodeWeightvsPosition}). Manual inspection of these early nodes shows that they primarily contain problem formalization and planning steps, including decomposing the task into sub-problems and deciding the overall reasoning strategy. In true negative predictions, importance shifts toward the middle of the trace. This pattern associates incorrect outcomes with intermediate reasoning structure; it does not establish that those nodes cause the error.

\paragraph{Planning and Synthesis Steps Are Crucial.}

We next examine how node connectivity (in-degree and out-degree)\footnote{In-degree and out-degree refer, respectively, to the number of incoming and outgoing edges of a node in a graph.} relates to \BLUE{final-answer correctness} (see Figure~\ref{fig:node_importance_degree}). High out-degree nodes (\emph{planner}) are typically early and correspond to planning and task decomposition in correct CoTs. High in-degree nodes (\emph{synthesis sinks}) are more associated with incorrect CoTs and correspond to steps where the LLM conditions the generation on multiple previous reasoning threads, background information, and verification loops.
Overall, correct reasoning shows initial planning and more distributed aggregation, while incorrect CoTs are characterized by dense synthesis steps.

\begin{figure}[!t]
	\centering
	\includegraphics[width=\linewidth]{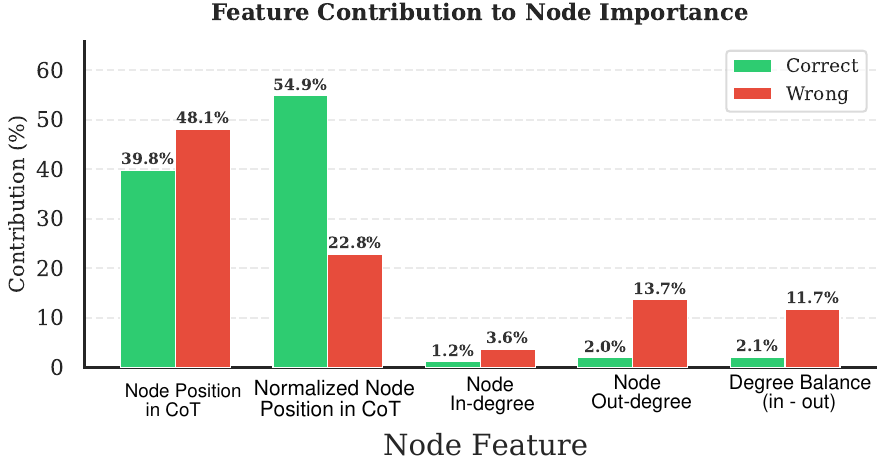}
	\caption{%
		Linear regression of node-feature importance scores, 
		comparing correct and incorrect CoTs.
	}
	\label{fig:feature_regression}
\end{figure}

\paragraph{Feature Attribution via Regression Analysis.}
A regression over GNNExplainer scores shows the same pattern: positional features dominate attribution for correct CoTs, whereas connectivity features receive relatively greater importance for incorrect traces (Figure~\ref{fig:feature_regression}). Normalized position accounts for 54.9\% of feature importance in correct CoTs and 22.8\% in incorrect CoTs, concentrating attribution on early planning in successful traces. Absolute position receives more importance in incorrect CoTs, which are typically longer~\cite{wu2026when}. Connectivity is more important in incorrect traces and is associated with subtask aggregation.

\subsection{Reasoning Efficiency Analysis}
\label{subsec:efficiency}

\ResultBox{Question}{Do long CoTs reflect useful exploration and verification, or structural inefficiency?}{Structural inefficiency dominates: it accounts for 49.0\% of steps on average, compared with 6.6\% for exploration/verification. Useful exploration emerges primarily in long CoTs (19.2\%).}

\begin{table*}[!t]
	\centering
	\small
	\setlength{\tabcolsep}{5pt}
	\renewcommand{\arraystretch}{1.2}
	\begin{tabular}{lcccccccc}
		\toprule
		\multirow{2}{*}{\textbf{Model}}
		 & \multicolumn{2}{c}{\textbf{Short} ($p0$--$33$)}
		 & \multicolumn{2}{c}{\textbf{Medium} ($p33$--$66$)}
		 & \multicolumn{2}{c}{\textbf{Long} ($p66$--$100$)}
		 & \multicolumn{2}{c}{\textbf{Overall}} \\
		\cmidrule(lr){2-3}\cmidrule(lr){4-5}\cmidrule(lr){6-7}\cmidrule(lr){8-9}
		 & $\mathcal{L}_{\mathrm{verif.}}$ & $\mathcal{L}_{\mathrm{ineff.}}$
		 & $\mathcal{L}_{\mathrm{verif.}}$ & $\mathcal{L}_{\mathrm{ineff.}}$
		 & $\mathcal{L}_{\mathrm{verif.}}$ & $\mathcal{L}_{\mathrm{ineff.}}$
		 & $\mathcal{L}_{\mathrm{verif.}}$ & $\mathcal{L}_{\mathrm{ineff.}}$ \\
		\midrule
		DS-Qw7B
		 & $0.0$ & $25.8$
		 & $2.2$ & $53.0$
		 & $20.4$ & $78.5$
		 & $7.3$ & $52.6$ \\
		DS-Qw14B
		 & $0.0$ & $9.7$
		 & $0.0$ & $34.1$
		 & $18.0$ & $61.8$
		 & $5.5$ & $35.5$ \\
		MiMo-7B-RL
		 & $0.0$ & $39.8$
		 & $2.2$ & $59.0$
		 & $19.2$ & $77.4$
		 & $6.9$ & $58.9$ \\
    \midrule
    \textbf{Average}
     & $0.0$ & $25.1$
     & $1.5$ & $48.7$
     & $19.2$ & $72.6$
     & $6.6$ & $49.0$ \\
    \bottomrule
	\end{tabular}
    \caption{%
		Decomposition of reasoning redundancy across CoT length on GPQA.
		\emph{$\mathcal{L}_{\mathrm{verif.}}$}\ = good redundancy (exploration/verification);
		\emph{$\mathcal{L}_{\mathrm{ineff.}}$}\ = bad redundancy (suboptimal order).
		Short/Medium/Long refer to CoT length quantile. 
	}
    \vspace{-0.8em}
   	\label{tab:redundancy_compact}
\end{table*}

\paragraph{Structural inefficiency dominates CoT redundancy.}
Most long-CoT redundancy reflects structural inefficiency rather than useful verification. Across models, path inefficiency accounts for 49.0\% of steps on average, compared with only 6.6\% attributable to verification/exploration. Useful redundancy is concentrated in long traces, reaching 19.2\%, whereas inefficiency is already present in short traces and increases with length (Table~\ref{tab:redundancy_compact}).
To distinguish useful exploration~\cite{wang2023selfconsistency} from unnecessary redundancy, we decompose each trace into two sources of efficiency loss: $(i)$ exploration/verification, where the model revisits or repeats reasoning, and $(ii)$ path inefficiency, where reasoning steps are removable or generated in a suboptimal dependency order. GPT-5 Mini relevance labels (\S\ref{subsec:labels}) identify removable steps, while graph edges define dependency constraints.

\paragraph{Structural Efficiency Decomposition.}

\begin{figure}[!t]
\centering
\begin{tikzpicture}[
  >={Stealth[length=1.6mm,width=1.3mm]},
  vtx/.style={circle, draw=black!60, fill=white, line width=0.55pt,
              minimum size=6.1mm, inner sep=0pt, font=\scriptsize},
  rdn/.style={vtx, fill=redamber!55, draw=redamber!85!black},
  mrg/.style={circle, draw=tspteal, fill=tspteal!10, line width=0.9pt,
              minimum size=8.0mm, inner sep=0pt, font=\tiny},
  term/.style={rectangle, rounded corners=2pt, draw=black!55, fill=black!6,
               minimum height=5.8mm, inner xsep=1.6mm, font=\scriptsize},
  dep/.style={->, draw=depgrey, line width=0.9pt},
  merged/.style={->, draw=tspteal!55, line width=0.9pt, densely dashed},
  seq/.style={circle, draw=black!55, fill=white, minimum size=3.2mm,
              inner sep=0pt, font=\tiny},
  sarr/.style={-{Stealth[length=0.85mm,width=0.7mm]}},
  seqr/.style={seq, fill=redamber!55, draw=redamber!85!black},
  seqm/.style={seq, draw=tspteal, fill=tspteal!12, line width=0.7pt},
  lbl/.style={font=\tiny\itshape, text=black!60},
]
\node[font=\scriptsize\bfseries, anchor=west] at (-0.25,1.60)
      {(a)~Example CoT graph after merging repeated steps};
\node[term] (P)  at (0.50,1.00)  {problem};
\node[vtx]  (v1) at (0.50,0)     {$v_1$};
\node[vtx]  (v2) at (2.18,0.78)  {$v_2$};
\node[vtx]  (v5) at (1.56,-0.62) {$v_5$};
\node[vtx]  (v6) at (2.96,-0.62) {$v_6$};
\node[mrg]  (v3) at (4.19,0.08)  {$v_3\!\equiv\! v_7$};
\node[rdn]  (v4) at (5.70,0.60)  {$v_4$};
\node[vtx]  (v8) at (4.92,-0.62) {$v_8$};
\node[term] (A)  at (6.13,-0.62) {answer};
\draw[dep] (P)--(v1); \draw[dep] (v1)--(v2); \draw[dep] (v1)--(v5);
\draw[dep] (v2)--(v3); \draw[dep] (v5)--(v6); \draw[merged] (v6)--(v3);
\draw[dep] (v3)--(v4); \draw[dep] (v3)--(v8); \draw[dep] (v6)--(v8);
\draw[dep] (v8)--(A);
\node[lbl, text=tspteal, anchor=south east, align=center] at (4.88,0.44)
      {Repetition:\\$v_7$ merged into $v_3$};
\draw[dep] (0.05,-1.12) -- (0.55,-1.12);
\node[anchor=west, font=\tiny, text=black!60] at (0.60,-1.12) {prerequisite};
\draw[merged] (2.20,-1.12) -- (2.70,-1.12);
\node[anchor=west, font=\tiny, text=black!60] at (2.75,-1.12) {redirected by merge};
\node[rdn, minimum size=3.2mm] at (5.05,-1.12) {};
\node[anchor=west, font=\tiny, text=black!60] at (5.25,-1.12) {judged redundant};

\node[font=\scriptsize\bfseries, anchor=north west] at (-0.25,-1.46)
      {(b)~Prerequisites};
\node[anchor=north west, font=\tiny] at (-0.25,-1.68) {%
\begin{tabular}{@{}l@{\hskip 5pt}l@{}}
\toprule
Node & Requires \\
\midrule
$v_1$              & problem \\
$v_2$              & $v_1$ \\
$v_5$              & $v_1$ \\
$v_6$              & $v_5$ \\
$v_3\!\equiv\!v_7$ & $v_2$,\; $v_6^{\dagger}$ \\
$v_4$              & $v_3$ \\
$v_8$              & $v_3$,\; $v_6$ \\
\bottomrule
\end{tabular}};
\node[anchor=north west, font=\tiny, text=black!60] at (-0.25,-4.18)
      {$^{\dagger}$ inherited from $v_7$};

\def\sx{3.50}
\def\sy{-2.24}
\def\dx{0.44}
\node[font=\scriptsize\bfseries, anchor=north west] at (2.42,-1.52)
      {(c)~Traversals};
\foreach \i/\lab/\st in {0/1/seq,1/2/seq,2/3/seqm,3/4/seqr,4/5/seq,5/6/seq,6/3/seqm,7/8/seq}{
  \node[\st] (c\i) at (\sx+\dx*\i,\sy) {\lab};}
\foreach \i in {0,...,6}{\pgfmathtruncatemacro{\j}{\i+1}
  \draw[sarr,cotgrey,line width=0.6pt] (c\i)--(c\j);}
\draw[->,cotgrey,line width=0.6pt,densely dotted] (c6) to[out=45,in=135] (c2);
\node[font=\tiny, text=cotgrey, anchor=south] at (\sx+\dx*4,\sy+0.50)
  {returns to $v_3$};
\node[anchor=east, font=\tiny, align=right] at (\sx-0.20,\sy)
      {$L_{\mathrm{CoT}}\!=\!8$};
\foreach \i/\lab/\st in {0/1/seq,1/2/seq,2/3/seqm,3/4/seqr,4/5/seq,5/6/seq,6/8/seq}{
  \node[\st] (l\i) at (\sx+\dx*\i,\sy-0.72) {\lab};}
\foreach \i in {0,...,5}{\pgfmathtruncatemacro{\j}{\i+1}
  \draw[sarr,loopblue,line width=0.7pt] (l\i)--(l\j);}
\node[anchor=east, font=\tiny, align=right] at (\sx-0.20,\sy-0.85)
      {$L_{\mathrm{loops}}\!=\!7$};
\foreach \i/\lab/\st in {0/1/seq,1/2/seq,2/5/seq,3/6/seq,4/3/seqm,5/8/seq}{
  \node[\st] (t\i) at (\sx+\dx*\i,\sy-1.44) {\lab};}
\foreach \i in {0,...,4}{\pgfmathtruncatemacro{\j}{\i+1}
  \draw[sarr,tspteal,line width=0.7pt,densely dashed] (t\i)--(t\j);}
\draw[tspteal!55, line width=0.55pt, rounded corners=1.2pt]
  ($(t2.north west)+(-0.45mm,0.45mm)$) rectangle ($(t4.south east)+(0.45mm,-0.45mm)$);
\node[anchor=east, font=\tiny, align=right] at (\sx-0.20,\sy-1.70)
      {$L_{\mathrm{TSP^*}}\!=\!6$};
\node[anchor=north, font=\tiny, text=tspteal] at (\sx+\dx*3,\sy-1.74)
      {$v_3$ deferred until $v_6$ is available};

\begin{scope}[on background layer]
  \draw[rounded corners=3pt, draw=black!35, fill=white, line width=0.6pt]
    ($(current bounding box.south west)+(-1.2mm,-0.7mm)$) rectangle
    ($(current bounding box.north east)+(1.2mm,1.1mm)$);
\end{scope}
\node[font=\small\bfseries, fill=white, inner xsep=3pt, inner ysep=1pt]
     at (current bounding box.north)
     {Reasoning Efficiency Analysis Visualization};
\end{tikzpicture}
\vspace{-0.35em}
\caption{Example of the efficiency decomposition. $v_7$ repeats $v_3$, so the two collapse into one node that inherits $v_7$'s prerequisite $v_6$; $v_4$ is judged redundant. $L_{\mathrm{TSP^*}}$ visits $v_5$ and $v_6$ before $v_3$, due to node prerequisites.}
\label{fig:efficiency_decomposition}
\end{figure}
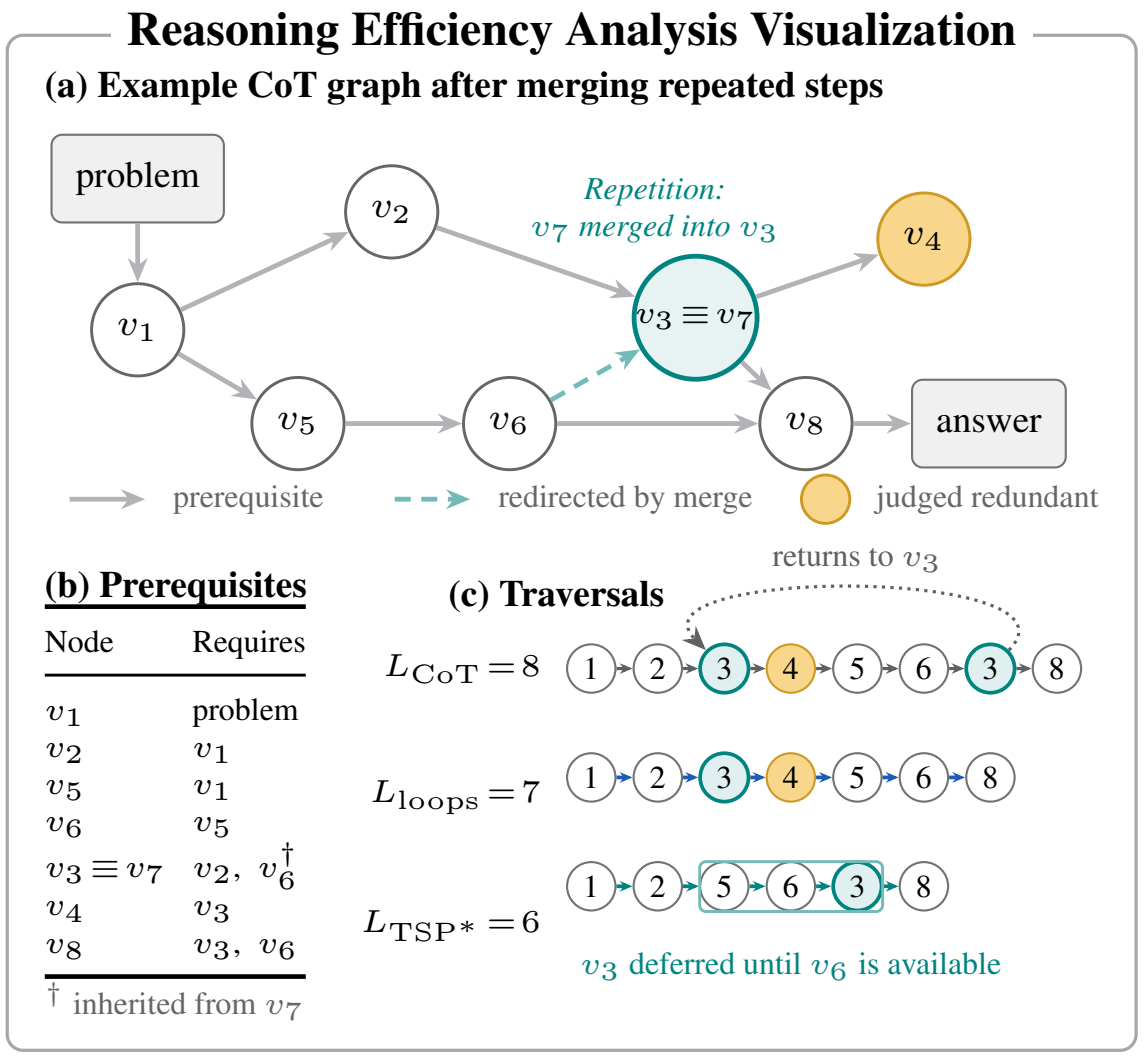

We quantify two sources of efficiency loss (Figure~\ref{fig:efficiency_decomposition}).
Let $L_{\mathrm{CoT}}$ be the original number of reasoning steps in the CoT, $L_{\mathrm{loops}}$ the trace length after collapsing repeated steps, and $L_{\mathrm{TSP^*}}$ the length of an approximate minimal traversal that respects the inferred graph dependencies. We then decompose total efficiency loss $\mathcal{L}$ into exploration/verification steps, $\mathcal{L}_{\mathrm{verif.}}$, and path inefficiency, $\mathcal{L}_{\mathrm{ineff.}}$:
\begin{equation}
\mathcal{L} = 
\underbrace{\frac{L_{\mathrm{CoT}} - L_{\mathrm{loops}}}{L_{\mathrm{CoT}}}}_{\mathcal{L}_{\mathrm{verif.}}}
+
\underbrace{\frac{L_{\mathrm{loops}} - L_{\mathrm{TSP^*}}}{L_{\mathrm{CoT}}}}_{\mathcal{L}_{\mathrm{ineff.}}}
\label{eq:efficiency}
\end{equation}

$\mathcal{L}_{\mathrm{verif.}}$ measures the steps removed by collapsing repetition loops, whereas $\mathcal{L}_{\mathrm{ineff.}}$ measures the additional reduction obtained by following a minimal dependency-consistent traversal.

\paragraph{Graph-Based Optimal Path Construction.}
\BLUE{$L_{\mathrm{TSP^*}}$ is the length of a traversal that visits a step only once all its prerequisites are available.} When no admissible next node exists, it relaxes the lowest-strength incoming prerequisite; equal-strength ties use the empirically derived type order. We use these prerequisite-relaxation events as the step-skipping signal for the hallucination analysis. Full details are given in Appx.~\ref{app:TSP}.

\begin{figure}[!t]
	\centering
    \text{Verification-Exploration Redundancy}
    \includegraphics[width=0.9\linewidth, trim=0 0 0 1.7em, clip]{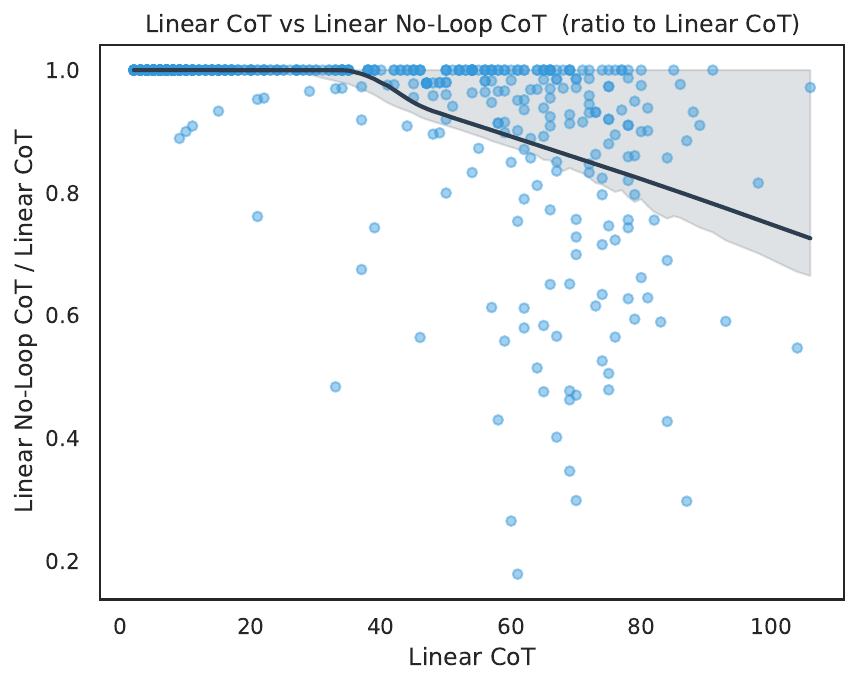}
    \caption{%
    Scatter plot of original CoT length ($L_{\mathrm{CoT}}$, x-axis) vs.\ efficiency
    ($E = 1 - \mathcal{L}_{\mathrm{verif.}}$, y-axis), where $\mathcal{L}_{\mathrm{verif.}}$ includes only good redundancy (verification/exploration) and excludes path inefficiency. We observe that verification loops start appearing
    once CoTs reach \BLUE{35--45} steps, and then efficiency decreases
    approximately linearly as CoT length increases.
    }
    \vspace{-0.8em}
	\label{fig:efficiency_noloop}
\end{figure}

\paragraph{Empirical Findings.}
Across models, path inefficiency accounts for 49.0\% of steps on average, against only 6.6\% for exploration/verification, and Table~\ref{tab:redundancy_compact} shows a clear
length-dependent pattern. Exploration/verification redundancy is negligible in short traces and appears primarily after approximately 35--45 steps (Figure~\ref{fig:efficiency_noloop}),
rising to 19.2\% in the longest traces.
Path inefficiency, in contrast, is already present in short traces and rises across all length quantiles. Among the evaluated models, DS-Qw14B exhibits substantially lower path inefficiency (35.5\%) than DS-Qw7B (52.6\%) and MiMo-7B-RL (58.9\%), despite comparable levels of verification redundancy.

\paragraph{Dependency Violations Are Associated with Step Errors.}
Complex reasoning often requires decomposing out-of-distribution problems into intermediate sub-tasks within the model's capabilities, followed by composition of partial solutions. In this setting, skipping intermediate steps can be a valid shortcut but may also lead to hallucinations. We next test whether generating a step before its inferred prerequisites are satisfied is associated with node-level errors. We define a skip as a prerequisite-relaxation event during the $L_{\mathrm{TSP^*}}$ traversal. Relative to correct nodes, partially hallucinated nodes exhibit +6.96\% higher skipping and hallucinated nodes +7.03\%, consistently across models (Figure~\ref{fig:hallucination_vs_stepskipping}).

\begin{figure}[!t]
	\centering
    \text{Best-of-$N$ Decoding}
	\includegraphics[width=\linewidth]{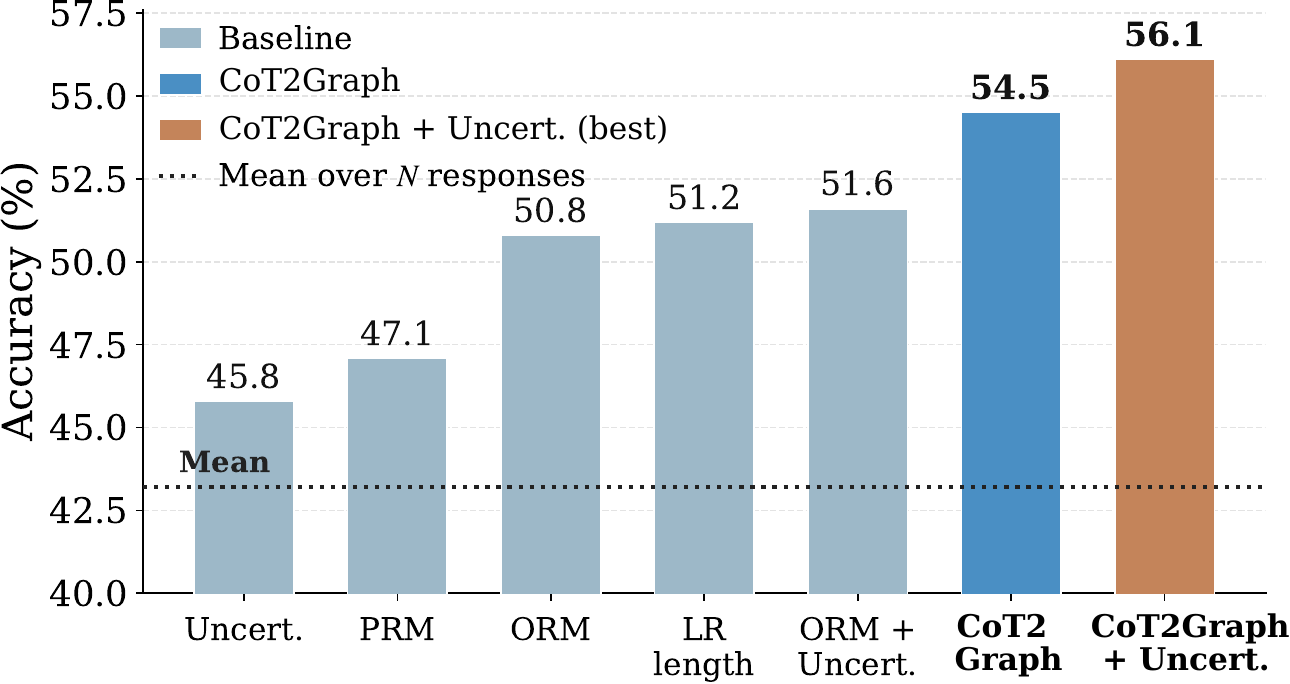}
	\caption{
        Best-of-$N$ ($N=5$) selection accuracy on MMLU-Pro with DS-Qw7B.  CoT2Graph-Best achieves the highest performance among single verifiers (54.5\%) with a baseline of 43.2\%, outperforming all previous methods by at least 2.9\%.  Combining CoT2Graph with entropy selection further improves accuracy to 56.1\%. 
	}
    \vspace{-0.8em}
	\label{fig:bestofN_comparison}
\end{figure}

\begin{figure*}[!t]
	\centering
	\includegraphics[width=\linewidth]{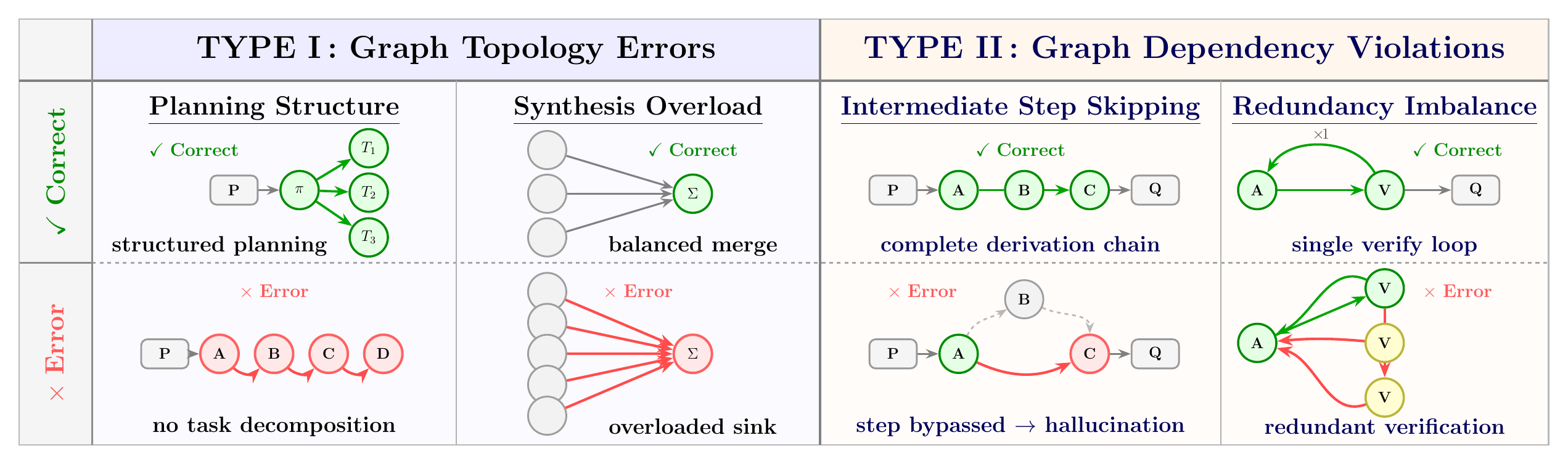}
	\caption{%
		Structural reasoning patterns identified by the proposed \textbf{CoT2Graph} taxonomy, including planning structures, synthesis node overload, intermediate step skipping, and verification redundancy.
	}
	\label{fig:pattern_error1}
\end{figure*}

\subsection{\BLUE{CoT2Graph Helps Best-of-\texorpdfstring{$N$}{N} Decoding}}
\label{subsec:bon}
\ResultBox{Question}{Can structural signals improve Best-of-\texorpdfstring{$N$}{N} selection, and are they complementary to uncertainty?}{CoT2Graph is the strongest standalone verifier; combining it with entropy improves accuracy to 56.1\%.}
We evaluate CoT2Graph as a \emph{structural verifier} in Best-of-$N$ sampling~\cite{deng-etal-2024-towards}. Given $N$ independent CoTs for the same question, we select the one most likely to yield a correct answer. Beyond absolute performance, we assess whether the methods capture complementary signals.

\paragraph{CoT Verifier Methods.}

We compare six selection strategies. The entropy method selects the CoT with the lowest mean per-token entropy~\cite{fu2025deepthinkconfidence}. PRM-Best (Mistral-7B-PRM) aggregates step-level rewards~\cite{lightman2024lets,wang2024mathshepherd}, while ORM-Best (Skywork-Reward-Llama-8B) uses a final outcome score~\cite{liu2026skyworkrewardv2scalingpreferencedata}. Logistic regression ranks CoTs by length, and CoT2Graph by GNN-predicted correctness. We also report ORM combined with entropy (ORM + Uncert.), the strongest non-graph baseline.
The PRM, ORM, and edge annotator have 7--8B parameters, but this does not make their costs or inputs equivalent: PRM and ORM score trace text, whereas the GNN receives graph-encoded structural features after edge annotation. Edge annotation considers at most $n(n-1)/2$ node pairs per trace.\footnote{See Appx.~\ref{app:cost_review} for details on runtime.}
We evaluate on 600 MMLU-Pro questions using DS-Qw7B and $N{=}5$ samples. The mean baseline achieves 43.2\%, while the oracle upper bound is 68.3\%. CoT2Graph-Best reaches 54.5\%, outperforming ORM-Best (50.8\%) and PRM-Best (47.1\%); the gain over the strongest non-graph verifier is 2.9 points (Figure~\ref{fig:bestofN_comparison}).

\paragraph{Orthogonal Signals and Ensemble Strategy.}
CoT2Graph is weakly correlated with entropy-based selection (\(\rho=-0.16\)), but moderately correlated with PRM, ORM, and length-based ranking (\(\rho>0.4\)) (see Figure~\ref{fig:bestofN_corr}). This suggests that topology and uncertainty capture complementary signals: combining CoT2Graph with entropy increases accuracy from 54.5\% to 56.1\%, the best overall result.


\section{Structural CoT Error Taxonomy}
\label{subsec:taxonomy}
Existing CoT evaluation frameworks, such as ROSCOE~\cite{golovneva2023roscoe}, mainly evaluate semantic properties (fluency, coherence, factual consistency, and logical validity) at the step level. They do not describe global structural patterns in the CoT (Figure~\ref{fig:pattern_error1}).
We organize the observed associations into two categories:

\paragraph{Type I: Graph Topology Errors (\S\ref{subsec:interpretation}).}
Topology patterns associated with reasoning outcome.
\begin{itemize}[leftmargin=*, nosep]
    \item \textbf{Planning structure}. Correct CoTs often contain early planning nodes with moderate out-degree, corresponding to task decomposition and branching. In contrast, incorrect CoTs more often follow linear chains without explicit planning.
    \item \textbf{Synthesis overload}. Correct reasoning tends to distribute synthesis operations across the trajectory, while incorrect CoTs contain high in-degree sink nodes that merge many dependencies at once. They mark reasoning failures and fragile synthesis steps.
\end{itemize}

\paragraph{Type II: Graph Dependency Violations (\S\ref{subsec:efficiency}).}
Patterns in which the observed step order conflicts with the inferred dependencies.
\begin{itemize}[leftmargin=*, nosep]
    \item \textbf{Intermediate step skipping}. Some CoTs bypass required intermediate derivations and produce conclusions before prerequisite subgoals are established, which raises hallucination rates.
    \item \textbf{Redundancy imbalance}. Moderate verification and exploration occur in correct traces, whereas excessive branching and repetition are associated with structural inefficiency and failure.
\end{itemize}

We also observe cases of structurally flawed reasoning with correct final answers, including traces that initially follow incorrect reasoning paths but self-correct near the final step, as well as multiple-choice reasoning by exclusion close to guessing.

\section{Conclusion}
We introduced CoT2Graph to study how the structure of CoT traces relates to reasoning quality. Representing long CoTs as typed dependency graphs improves correctness prediction over tree- and length-based baselines and enables interpretable analysis of reasoning patterns. GNNExplainer reveals that correct CoTs emphasize early planning and task decomposition, whereas incorrect CoTs are associated with overloaded synthesis steps and dependency violations. We further find that much of long-CoT redundancy reflects structural inefficiency rather than useful exploration or verification, and that structural signals complement uncertainty-based Best-of-\texorpdfstring{$N$}{N} selection.
Overall, our findings suggest that reasoning quality is reflected not only in individual steps, but also in how those steps are organized and composed, motivating moving from sequence- and step-level evaluation toward structure-aware analysis of reasoning traces.

\section*{Limitations}

In this work, we employ an LLM-as-Judge setup for relevance and hallucination annotations, which may inherit biases from the GPT-5 Mini evaluator~\cite{singh2026openaigpt5card}. While this provides flexible and scalable supervision, alternative annotation strategies could rely on smaller NLI models or task-specific classifiers for edge and node labeling.

The quality of the resulting graphs is also influenced by the accuracy of the underlying step segmentation process, which determines the granularity of CoT decomposition and therefore the structure of the constructed graph.

Edge labels are generated by Llama-3.1-8B-Instruct and remain subject to judge and prompt bias. Appx.~\ref{app:annotation_validation_review} reports the 95-edge human comparison, where judge--ensemble agreement is \BLUE{85.3\%} and Cohen $\kappa$ is \BLUE{0.82}. Pairwise annotation also limits use on long traces.

Overall, the goal of this framework is to analyze and expose structural signals in Chain-of-Thought reasoning traces, rather than to define a single optimal annotation or construction paradigm.

\section*{Ethical considerations}

We do not identify specific ethical risks associated with our work. The framework supports interpretability by describing reasoning patterns associated with failures and hallucinations.

We believe that improving the transparency and analysis of reasoning processes can contribute to the development of more reliable and safer AI systems, particularly as LLMs are increasingly deployed in autonomous and agentic settings.

\section*{Acknowledgments}

The authors thank the reviewers for their constructive feedback. T.~F.~Banfi is currently affiliated with CERN; this work was conducted before, while he was at ETH Zurich (LRE Lab).
M.~Akhtar was supported by the ETH AI Center through an ETH AI Center postdoctoral fellowship.

We thank Stefano Marin (CERN) and Noah Julian St\"auble (CERN) for their thoughtful comments and feedback in the late stage of this work.

We also thank Francesco Dorati (Politecnico di Milano) for his help and time dedicated to the human validation and reliability analysis of the LLM-as-a-Judge edge annotations.

\bibliography{custom}

@inproceedings{jiang-etal-2025-makes,
  title     = {What Makes a Good Reasoning Chain? Uncovering Structural Patterns in Long Chain-of-Thought Reasoning},
  author    = {Jiang, Gangwei  and
               Liu, Yahui  and
               Li, Zhaoyi  and
               Bi, Wei  and
               Zhang, Fuzheng  and
               Song, Linqi  and
               Wei, Ying  and
               Lian, Defu},
  editor    = {Christodoulopoulos, Christos  and
               Chakraborty, Tanmoy  and
               Rose, Carolyn  and
               Peng, Violet},
  booktitle = {Proceedings of the 2025 Conference on Empirical Methods in Natural Language Processing},
  month     = nov,
  year      = {2025},
  address   = {Suzhou, China},
  publisher = {Association for Computational Linguistics},
  url       = {https://aclanthology.org/2025.emnlp-main.329/},
  doi       = {10.18653/v1/2025.emnlp-main.329},
  pages     = {6490--6514},
  isbn      = {979-8-89176-332-6}
}

@inproceedings{prasad-etal-2023-receval,
  title     = {{R}e{CE}val: Evaluating Reasoning Chains via Correctness and Informativeness},
  author    = {Prasad, Archiki  and
               Saha, Swarnadeep  and
               Zhou, Xiang  and
               Bansal, Mohit},
  editor    = {Bouamor, Houda  and
               Pino, Juan  and
               Bali, Kalika},
  booktitle = {Proceedings of the 2023 Conference on Empirical Methods in Natural Language Processing},
  month     = dec,
  year      = {2023},
  address   = {Singapore},
  publisher = {Association for Computational Linguistics},
  url       = {https://aclanthology.org/2023.emnlp-main.622/},
  doi       = {10.18653/v1/2023.emnlp-main.622},
  pages     = {10066--10086}
}

@inproceedings{golovneva2023roscoe,
  title     = {{ROSCOE}: A Suite of Metrics for Scoring Step-by-Step Reasoning},
  author    = {Olga Golovneva and Moya Peng Chen and Spencer Poff and Martin Corredor and Luke Zettlemoyer and Maryam Fazel-Zarandi and Asli Celikyilmaz},
  booktitle = {The Eleventh International Conference on Learning Representations},
  year      = {2023},
  url       = {https://openreview.net/forum?id=xYlJRpzZtsY}
}

@inproceedings{mishra-etal-2025-investigating,
  title     = {Investigating the Shortcomings of {LLM}s in Step-by-Step Legal Reasoning},
  author    = {Mishra, Venkatesh  and
               Pathiraja, Bimsara  and
               Parmar, Mihir  and
               Chidananda, Sat  and
               Srinivasa, Jayanth  and
               Liu, Gaowen  and
               Payani, Ali  and
               Baral, Chitta},
  editor    = {Chiruzzo, Luis  and
               Ritter, Alan  and
               Wang, Lu},
  booktitle = {Findings of the Association for Computational Linguistics: NAACL 2025},
  month     = apr,
  year      = {2025},
  address   = {Albuquerque, New Mexico},
  publisher = {Association for Computational Linguistics},
  url       = {https://aclanthology.org/2025.findings-naacl.435/},
  doi       = {10.18653/v1/2025.findings-naacl.435},
  pages     = {7810--7841},
  isbn      = {979-8-89176-195-7}
}

@inproceedings{deng-etal-2024-towards,
  title     = {Towards A Unified View of Answer Calibration for Multi-Step Reasoning},
  author    = {Deng, Shumin  and
               Zhang, Ningyu  and
               Oo, Nay  and
               Hooi, Bryan},
  editor    = {Dalvi Mishra, Bhavana  and
               Durrett, Greg  and
               Jansen, Peter  and
               Lipkin, Ben  and
               Neves Ribeiro, Danilo  and
               Wong, Lionel  and
               Ye, Xi  and
               Zhao, Wenting},
  booktitle = {Proceedings of the 2nd Workshop on Natural Language Reasoning and Structured Explanations (@ACL 2024)},
  month     = aug,
  year      = {2024},
  address   = {Bangkok, Thailand},
  publisher = {Association for Computational Linguistics},
  url       = {https://aclanthology.org/2024.nlrse-1.3/},
  pages     = {25--38}
}

@inproceedings{ijcai2025p925,
  title     = {Can We Verify Step by Step for Incorrect Answer Detection?},
  author    = {Xu, Xin and Diao, Shizhe and Yang, Can and Wang, Yang},
  booktitle = {Proceedings of the Thirty-Fourth International Joint Conference on
               Artificial Intelligence, {IJCAI-25}},
  publisher = {International Joint Conferences on Artificial Intelligence Organization},
  editor    = {James Kwok},
  pages     = {8322--8330},
  year      = {2025},
  month     = {8},
  note      = {Main Track},
  doi       = {10.24963/ijcai.2025/925},
  url       = {https://doi.org/10.24963/ijcai.2025/925},
}

@inproceedings{li-etal-2024-evaluating-mathematical,
  title     = {Evaluating Mathematical Reasoning of Large Language Models: A Focus on Error Identification and Correction},
  author    = {Li, Xiaoyuan  and
               Wang, Wenjie  and
               Li, Moxin  and
               Guo, Junrong  and
               Zhang, Yang  and
               Feng, Fuli},
  editor    = {Ku, Lun-Wei  and
               Martins, Andre  and
               Srikumar, Vivek},
  booktitle = {Findings of the Association for Computational Linguistics: ACL 2024},
  month     = aug,
  year      = {2024},
  address   = {Bangkok, Thailand},
  publisher = {Association for Computational Linguistics},
  url       = {https://aclanthology.org/2024.findings-acl.673/},
  doi       = {10.18653/v1/2024.findings-acl.673},
  pages     = {11316--11360}
}

@inproceedings{jacovi2024chain,
  title     = {A chain-of-thought is as strong as its weakest link: A benchmark for verifiers of reasoning chains},
  author    = {Jacovi, Alon and Bitton, Yonatan and Bohnet, Bernd and Herzig, Jonathan and Honovich, Or and Tseng, Michael and Collins, Michael and Aharoni, Roee and Geva, Mor},
  booktitle = {Proceedings of the 62nd Annual Meeting of the Association for Computational Linguistics (Volume 1: Long Papers)},
  pages     = {4615--4634},
  year      = {2024}
}

@misc{xue2023rcotdetectingrectifyingfactual,
  title         = {RCOT: Detecting and Rectifying Factual Inconsistency in Reasoning by Reversing Chain-of-Thought},
  author        = {Tianci Xue and Ziqi Wang and Zhenhailong Wang and Chi Han and Pengfei Yu and Heng Ji},
  year          = {2023},
  eprint        = {2305.11499},
  archiveprefix = {arXiv},
  primaryclass  = {cs.CL},
  url           = {https://arxiv.org/abs/2305.11499}
}

@inproceedings{zhao-etal-2023-verify,
  title     = {Verify-and-Edit: A Knowledge-Enhanced Chain-of-Thought Framework},
  author    = {Zhao, Ruochen  and
               Li, Xingxuan  and
               Joty, Shafiq  and
               Qin, Chengwei  and
               Bing, Lidong},
  editor    = {Rogers, Anna  and
               Boyd-Graber, Jordan  and
               Okazaki, Naoaki},
  booktitle = {Proceedings of the 61st Annual Meeting of the Association for Computational Linguistics (Volume 1: Long Papers)},
  month     = jul,
  year      = {2023},
  address   = {Toronto, Canada},
  publisher = {Association for Computational Linguistics},
  url       = {https://aclanthology.org/2023.acl-long.320/},
  doi       = {10.18653/v1/2023.acl-long.320},
  pages     = {5823--5840}
}

@inproceedings{NEURIPS2024_01025a4e,
  author    = {Xue, Shangzi and Huang, Zhenya and Liu, Jiayu and lin, Xin and Ning, Yuting and Jin, Binbin and Li, Xin and Liu, Qi},
  booktitle = {Advances in Neural Information Processing Systems},
  doi       = {10.52202/079017-0012},
  editor    = {A. Globerson and L. Mackey and D. Belgrave and A. Fan and U. Paquet and J. Tomczak and C. Zhang},
  pages     = {357--385},
  publisher = {Curran Associates, Inc.},
  title     = {Decompose, Analyze and Rethink: Solving Intricate Problems with Human-like Reasoning Cycle},
  url       = {https://proceedings.neurips.cc/paper_files/paper/2024/file/01025a4e79355bb37a10ba39605944b5-Paper-Conference.pdf},
  volume    = {37},
  year      = {2024}
}

@inproceedings{jiang2025mmecot,
  title     = {{MME}-{CoT}: Benchmarking Chain-of-Thought in Large Multimodal Models for Reasoning Quality, Robustness, and Efficiency},
  author    = {Dongzhi Jiang and Renrui Zhang and Ziyu Guo and Yanwei Li and Yu Qi and Xinyan Chen and Liuhui Wang and Jianhan Jin and Claire Guo and Shen Yan and Bo Zhang and Chaoyou Fu and Peng Gao and Hongsheng Li},
  booktitle = {Forty-second International Conference on Machine Learning},
  year      = {2025},
  url       = {https://openreview.net/forum?id=YZvefQVLJI}
}

@inproceedings{lee-hockenmaier-2025-evaluating,
  title     = {Evaluating Step-by-step Reasoning Traces: A Survey},
  author    = {Lee, Jinu  and
               Hockenmaier, Julia},
  editor    = {Christodoulopoulos, Christos  and
               Chakraborty, Tanmoy  and
               Rose, Carolyn  and
               Peng, Violet},
  booktitle = {Findings of the Association for Computational Linguistics: EMNLP 2025},
  month     = nov,
  year      = {2025},
  address   = {Suzhou, China},
  publisher = {Association for Computational Linguistics},
  url       = {https://aclanthology.org/2025.findings-emnlp.94/},
  doi       = {10.18653/v1/2025.findings-emnlp.94},
  pages     = {1789--1814},
  isbn      = {979-8-89176-335-7}
}

@inproceedings{10.5555/3666122.3666639,
  author    = {Yao, Shunyu and Yu, Dian and Zhao, Jeffrey and Shafran, Izhak and Griffiths, Thomas L. and Cao, Yuan and Narasimhan, Karthik},
  title     = {Tree of thoughts: deliberate problem solving with large language models},
  year      = {2023},
  publisher = {Curran Associates Inc.},
  address   = {Red Hook, NY, USA},
  booktitle = {Proceedings of the 37th International Conference on Neural Information Processing Systems},
  articleno = {517},
  numpages  = {14},
  location  = {New Orleans, LA, USA},
  series    = {NIPS '23}
}

@inproceedings{10.1609/aaai.v38i16.29720,
  author    = {Besta, Maciej and Blach, Nils and Kubicek, Ales and Gerstenberger, Robert and Podstawski, Micha\l{} and Gianinazzi, Lukas and Gajda, Joanna and Lehmann, Tomasz and Niewiadomski, Hubert and Nyczyk, Piotr and Hoefler, Torsten},
  title     = {Graph of thoughts: solving elaborate problems with large language models},
  year      = {2024},
  isbn      = {978-1-57735-887-9},
  publisher = {AAAI Press},
  url       = {https://doi.org/10.1609/aaai.v38i16.29720},
  doi       = {10.1609/aaai.v38i16.29720},
  booktitle = {Proceedings of the Thirty-Eighth AAAI Conference on Artificial Intelligence and Thirty-Sixth Conference on Innovative Applications of Artificial Intelligence and Fourteenth Symposium on Educational Advances in Artificial Intelligence},
  articleno = {1972},
  numpages  = {9},
  series    = {AAAI'24/IAAI'24/EAAI'24}
}

@inproceedings{dalvi-etal-2021-explaining,
  title     = {Explaining Answers with Entailment Trees},
  author    = {Dalvi, Bhavana  and
               Jansen, Peter  and
               Tafjord, Oyvind  and
               Xie, Zhengnan  and
               Smith, Hannah  and
               Pipatanangkura, Leighanna  and
               Clark, Peter},
  editor    = {Moens, Marie-Francine  and
               Huang, Xuanjing  and
               Specia, Lucia  and
               Yih, Scott Wen-tau},
  booktitle = {Proceedings of the 2021 Conference on Empirical Methods in Natural Language Processing},
  month     = nov,
  year      = {2021},
  address   = {Online and Punta Cana, Dominican Republic},
  publisher = {Association for Computational Linguistics},
  url       = {https://aclanthology.org/2021.emnlp-main.585/},
  doi       = {10.18653/v1/2021.emnlp-main.585},
  pages     = {7358--7370}
}

@misc{fu2025deepthinkconfidence,
  title         = {Deep Think with Confidence},
  author        = {Yichao Fu and Xuewei Wang and Yuandong Tian and Jiawei Zhao},
  year          = {2025},
  eprint        = {2508.15260},
  archiveprefix = {arXiv},
  primaryclass  = {cs.LG},
  url           = {https://arxiv.org/abs/2508.15260}
}

@misc{ni2026reprobeefficienttesttimescaling,
  title         = {ReProbe: Efficient Test-Time Scaling of Multi-Step Reasoning by Probing Internal States of Large Language Models},
  author        = {Jingwei Ni and Ekaterina Fadeeva and Tianyi Wu and Mubashara Akhtar and Jiaheng Zhang and Elliott Ash and Markus Leippold and Timothy Baldwin and See-Kiong Ng and Artem Shelmanov and Mrinmaya Sachan},
  year          = {2026},
  eprint        = {2511.06209},
  archiveprefix = {arXiv},
  primaryclass  = {cs.AI},
  url           = {https://arxiv.org/abs/2511.06209}
}

@misc{banfi2026llmsgametheoryentropyguided,
  title         = {LLMs for Game Theory: Entropy-Guided In-Context Learning and Adaptive CoT Reasoning},
  author        = {Tommaso Felice Banfi and Sashenka Gamage},
  year          = {2026},
  eprint        = {2601.10775},
  archiveprefix = {arXiv},
  primaryclass  = {cs.CL},
  url           = {https://arxiv.org/abs/2601.10775}
}

@inproceedings{
wu2026when,
title={When More is Less: Understanding Chain-of-Thought Length in {LLM}s},
author={Yuyang Wu and Yifei Wang and Ziyu Ye and Tianqi Du and Stefanie Jegelka and Yisen Wang},
booktitle={The Fourteenth International Conference on Learning Representations},
year={2026},
url={https://openreview.net/forum?id=6QDFsYxtI1}
}

@inproceedings{wang2023selfconsistency,
  title     = {Self-Consistency Improves Chain of Thought Reasoning in Language Models},
  author    = {Xuezhi Wang and Jason Wei and Dale Schuurmans and Quoc V Le and Ed H. Chi and Sharan Narang and Aakanksha Chowdhery and Denny Zhou},
  booktitle = {The Eleventh International Conference on Learning Representations},
  year      = {2023},
  url       = {https://openreview.net/forum?id=1PL1NIMMrw}
}

@article{Guo_2025,
  title     = {DeepSeek-R1 incentivizes reasoning in LLMs through reinforcement learning},
  volume    = {645},
  issn      = {1476-4687},
  url       = {http://dx.doi.org/10.1038/s41586-025-09422-z},
  doi       = {10.1038/s41586-025-09422-z},
  number    = {8081},
  journal   = {Nature},
  publisher = {Springer Science and Business Media LLC},
  author    = {Guo, Daya and Yang, Dejian and Zhang, Haowei and Song, Junxiao and Wang, Peiyi and Zhu, Qihao and Xu, Runxin and Zhang, Ruoyu and Ma, Shirong and Bi, Xiao and Zhang, Xiaokang and Yu, Xingkai and Wu, Yu and Wu, Z. F. and Gou, Zhibin and Shao, Zhihong and Li, Zhuoshu and Gao, Ziyi and Liu, Aixin and Xue, Bing and Wang, Bingxuan and Wu, Bochao and Feng, Bei and Lu, Chengda and Zhao, Chenggang and Deng, Chengqi and Ruan, Chong and Dai, Damai and Chen, Deli and Ji, Dongjie and Li, Erhang and Lin, Fangyun and Dai, Fucong and Luo, Fuli and Hao, Guangbo and Chen, Guanting and Li, Guowei and Zhang, H. and Xu, Hanwei and Ding, Honghui and Gao, Huazuo and Qu, Hui and Li, Hui and Guo, Jianzhong and Li, Jiashi and Chen, Jingchang and Yuan, Jingyang and Tu, Jinhao and Qiu, Junjie and Li, Junlong and Cai, J. L. and Ni, Jiaqi and Liang, Jian and Chen, Jin and Dong, Kai and Hu, Kai and You, Kaichao and Gao, Kaige and Guan, Kang and Huang, Kexin and Yu, Kuai and Wang, Lean and Zhang, Lecong and Zhao, Liang and Wang, Litong and Zhang, Liyue and Xu, Lei and Xia, Leyi and Zhang, Mingchuan and Zhang, Minghua and Tang, Minghui and Zhou, Mingxu and Li, Meng and Wang, Miaojun and Li, Mingming and Tian, Ning and Huang, Panpan and Zhang, Peng and Wang, Qiancheng and Chen, Qinyu and Du, Qiushi and Ge, Ruiqi and Zhang, Ruisong and Pan, Ruizhe and Wang, Runji and Chen, R. J. and Jin, R. L. and Chen, Ruyi and Lu, Shanghao and Zhou, Shangyan and Chen, Shanhuang and Ye, Shengfeng and Wang, Shiyu and Yu, Shuiping and Zhou, Shunfeng and Pan, Shuting and Li, S. S. and Zhou, Shuang and Wu, Shaoqing and Yun, Tao and Pei, Tian and Sun, Tianyu and Wang, T. and Zeng, Wangding and Liu, Wen and Liang, Wenfeng and Gao, Wenjun and Yu, Wenqin and Zhang, Wentao and Xiao, W. L. and An, Wei and Liu, Xiaodong and Wang, Xiaohan and Chen, Xiaokang and Nie, Xiaotao and Cheng, Xin and Liu, Xin and Xie, Xin and Liu, Xingchao and Yang, Xinyu and Li, Xinyuan and Su, Xuecheng and Lin, Xuheng and Li, X. Q. and Jin, Xiangyue and Shen, Xiaojin and Chen, Xiaosha and Sun, Xiaowen and Wang, Xiaoxiang and Song, Xinnan and Zhou, Xinyi and Wang, Xianzu and Shan, Xinxia and Li, Y. K. and Wang, Y. Q. and Wei, Y. X. and Zhang, Yang and Xu, Yanhong and Li, Yao and Zhao, Yao and Sun, Yaofeng and Wang, Yaohui and Yu, Yi and Zhang, Yichao and Shi, Yifan and Xiong, Yiliang and He, Ying and Piao, Yishi and Wang, Yisong and Tan, Yixuan and Ma, Yiyang and Liu, Yiyuan and Guo, Yongqiang and Ou, Yuan and Wang, Yuduan and Gong, Yue and Zou, Yuheng and He, Yujia and Xiong, Yunfan and Luo, Yuxiang and You, Yuxiang and Liu, Yuxuan and Zhou, Yuyang and Zhu, Y. X. and Huang, Yanping and Li, Yaohui and Zheng, Yi and Zhu, Yuchen and Ma, Yunxian and Tang, Ying and Zha, Yukun and Yan, Yuting and Ren, Z. Z. and Ren, Zehui and Sha, Zhangli and Fu, Zhe and Xu, Zhean and Xie, Zhenda and Zhang, Zhengyan and Hao, Zhewen and Ma, Zhicheng and Yan, Zhigang and Wu, Zhiyu and Gu, Zihui and Zhu, Zijia and Liu, Zijun and Li, Zilin and Xie, Ziwei and Song, Ziyang and Pan, Zizheng and Huang, Zhen and Xu, Zhipeng and Zhang, Zhongyu and Zhang, Zhen},
  year      = {2025},
  month     = sep,
  pages     = {633--638}
}

@inproceedings{
chen2025research,
title={ReSearch: Learning to Reason with Search for {LLM}s via Reinforcement Learning},
author={Mingyang Chen and Linzhuang Sun and Tianpeng Li and Haoze Sun and Yijie Zhou and Chenzheng Zhu and Haofen Wang and Jeff Z. Pan and Wen Zhang and Huajun Chen and Fan Yang and Zenan Zhou and Weipeng Chen},
booktitle={The Thirty-ninth Annual Conference on Neural Information Processing Systems},
year={2025},
url={https://openreview.net/forum?id=OuGAwwAT8G}
}

@misc{singh2026openaigpt5card,
  title         = {OpenAI GPT-5 System Card},
  author        = {Aaditya Singh and Adam Fry and Adam Perelman and Adam Tart and Adi Ganesh and Ahmed El-Kishky and Aidan McLaughlin and Aiden Low and AJ Ostrow and Akhila Ananthram and Akshay Nathan and Alan Luo and Alec Helyar and Aleksander Madry and Aleksandr Efremov and Aleksandra Spyra and Alex Baker-Whitcomb and Alex Beutel and Alex Karpenko and Alex Makelov and Alex Neitz and Alex Wei and Alexandra Barr and Alexandre Kirchmeyer and Alexey Ivanov and Alexi Christakis and Alistair Gillespie and Allison Tam and Ally Bennett and Alvin Wan and Alyssa Huang and Amy McDonald Sandjideh and Amy Yang and Ananya Kumar and Andre Saraiva and Andrea Vallone and Andrei Gheorghe and Andres Garcia Garcia and Andrew Braunstein and Andrew Liu and Andrew Schmidt and Andrey Mereskin and Andrey Mishchenko and Andy Applebaum and Andy Rogerson and Ann Rajan and Annie Wei and Anoop Kotha and Anubha Srivastava and Anushree Agrawal and Arun Vijayvergiya and Ashley Tyra and Ashvin Nair and Avi Nayak and Ben Eggers and Bessie Ji and Beth Hoover and Bill Chen and Blair Chen and Boaz Barak and Borys Minaiev and Botao Hao and Bowen Baker and Brad Lightcap and Brandon McKinzie and Brandon Wang and Brendan Quinn and Brian Fioca and Brian Hsu and Brian Yang and Brian Yu and Brian Zhang and Brittany Brenner and Callie Riggins Zetino and Cameron Raymond and Camillo Lugaresi and Carolina Paz and Cary Hudson and Cedric Whitney and Chak Li and Charles Chen and Charlotte Cole and Chelsea Voss and Chen Ding and Chen Shen and Chengdu Huang and Chris Colby and Chris Hallacy and Chris Koch and Chris Lu and Christina Kaplan and Christina Kim and CJ Minott-Henriques and Cliff Frey and Cody Yu and Coley Czarnecki and Colin Reid and Colin Wei and Cory Decareaux and Cristina Scheau and Cyril Zhang and Cyrus Forbes and Da Tang and Dakota Goldberg and Dan Roberts and Dana Palmie and Daniel Kappler and Daniel Levine and Daniel Wright and Dave Leo and David Lin and David Robinson and Declan Grabb and Derek Chen and Derek Lim and Derek Salama and Dibya Bhattacharjee and Dimitris Tsipras and Dinghua Li and Dingli Yu and DJ Strouse and Drew Williams and Dylan Hunn and Ed Bayes and Edwin Arbus and Ekin Akyurek and Elaine Ya Le and Elana Widmann and Eli Yani and Elizabeth Proehl and Enis Sert and Enoch Cheung and Eri Schwartz and Eric Han and Eric Jiang and Eric Mitchell and Eric Sigler and Eric Wallace and Erik Ritter and Erin Kavanaugh and Evan Mays and Evgenii Nikishin and Fangyuan Li and Felipe Petroski Such and Filipe de Avila Belbute Peres and Filippo Raso and Florent Bekerman and Foivos Tsimpourlas and Fotis Chantzis and Francis Song and Francis Zhang and Gaby Raila and Garrett McGrath and Gary Briggs and Gary Yang and Giambattista Parascandolo and Gildas Chabot and Grace Kim and Grace Zhao and Gregory Valiant and Guillaume Leclerc and Hadi Salman and Hanson Wang and Hao Sheng and Haoming Jiang and Haoyu Wang and Haozhun Jin and Harshit Sikchi and Heather Schmidt and Henry Aspegren and Honglin Chen and Huida Qiu and Hunter Lightman and Ian Covert and Ian Kivlichan and Ian Silber and Ian Sohl and Ibrahim Hammoud and Ignasi Clavera and Ikai Lan and Ilge Akkaya and Ilya Kostrikov and Irina Kofman and Isak Etinger and Ishaan Singal and Jackie Hehir and Jacob Huh and Jacqueline Pan and Jake Wilczynski and Jakub Pachocki and James Lee and James Quinn and Jamie Kiros and Janvi Kalra and Jasmyn Samaroo and Jason Wang and Jason Wolfe and Jay Chen and Jay Wang and Jean Harb and Jeffrey Han and Jeffrey Wang and Jennifer Zhao and Jeremy Chen and Jerene Yang and Jerry Tworek and Jesse Chand and Jessica Landon and Jessica Liang and Ji Lin and Jiancheng Liu and Jianfeng Wang and Jie Tang and Jihan Yin and Joanne Jang and Joel Morris and Joey Flynn and Johannes Ferstad and Johannes Heidecke and John Fishbein and John Hallman and Jonah Grant and Jonathan Chien and Jonathan Gordon and Jongsoo Park and Jordan Liss and Jos Kraaijeveld and Joseph Guay and Joseph Mo and Josh Lawson and Josh McGrath and Joshua Vendrow and Joy Jiao and Julian Lee and Julie Steele and Julie Wang and Junhua Mao and Kai Chen and Kai Hayashi and Kai Xiao and Kamyar Salahi and Kan Wu and Karan Sekhri and Karan Sharma and Karan Singhal and Karen Li and Kenny Nguyen and Keren Gu-Lemberg and Kevin King and Kevin Liu and Kevin Stone and Kevin Yu and Kristen Ying and Kristian Georgiev and Kristie Lim and Kushal Tirumala and Kyle Miller and Lama Ahmad and Larry Lv and Laura Clare and Laurance Fauconnet and Lauren Itow and Lauren Yang and Laurentia Romaniuk and Leah Anise and Lee Byron and Leher Pathak and Leon Maksin and Leyan Lo and Leyton Ho and Li Jing and Liang Wu and Liang Xiong and Lien Mamitsuka and Lin Yang and Lindsay McCallum and Lindsey Held and Liz Bourgeois and Logan Engstrom and Lorenz Kuhn and Louis Feuvrier and Lu Zhang and Lucas Switzer and Lukas Kondraciuk and Lukasz Kaiser and Manas Joglekar and Mandeep Singh and Mandip Shah and Manuka Stratta and Marcus Williams and Mark Chen and Mark Sun and Marselus Cayton and Martin Li and Marvin Zhang and Marwan Aljubeh and Matt Nichols and Matthew Haines and Max Schwarzer and Mayank Gupta and Meghan Shah and Melody Y. Guan and Melody Huang and Meng Dong and Mengqing Wang and Mia Glaese and Micah Carroll and Michael Lampe and Michael Malek and Michael Sharman and Michael Zhang and Michele Wang and Michelle Pokrass and Mihai Florian and Mikhail Pavlov and Miles Wang and Ming Chen and Mingxuan Wang and Minnia Feng and Mo Bavarian and Molly Lin and Moose Abdool and Mostafa Rohaninejad and Nacho Soto and Natalie Staudacher and Natan LaFontaine and Nathan Marwell and Nelson Liu and Nick Preston and Nick Turley and Nicklas Ansman and Nicole Blades and Nikil Pancha and Nikita Mikhaylin and Niko Felix and Nikunj Handa and Nishant Rai and Nitish Keskar and Noam Brown and Ofir Nachum and Oleg Boiko and Oleg Murk and Olivia Watkins and Oona Gleeson and Pamela Mishkin and Patryk Lesiewicz and Paul Baltescu and Pavel Belov and Peter Zhokhov and Philip Pronin and Phillip Guo and Phoebe Thacker and Qi Liu and Qiming Yuan and Qinghua Liu and Rachel Dias and Rachel Puckett and Rahul Arora and Ravi Teja Mullapudi and Raz Gaon and Reah Miyara and Rennie Song and Rishabh Aggarwal and RJ Marsan and Robel Yemiru and Robert Xiong and Rohan Kshirsagar and Rohan Nuttall and Roman Tsiupa and Ronen Eldan and Rose Wang and Roshan James and Roy Ziv and Rui Shu and Ruslan Nigmatullin and Saachi Jain and Saam Talaie and Sam Altman and Sam Arnesen and Sam Toizer and Sam Toyer and Samuel Miserendino and Sandhini Agarwal and Sarah Yoo and Savannah Heon and Scott Ethersmith and Sean Grove and Sean Taylor and Sebastien Bubeck and Sever Banesiu and Shaokyi Amdo and Shengjia Zhao and Sherwin Wu and Shibani Santurkar and Shiyu Zhao and Shraman Ray Chaudhuri and Shreyas Krishnaswamy and Shuaiqi and Xia and Shuyang Cheng and Shyamal Anadkat and Simón Posada Fishman and Simon Tobin and Siyuan Fu and Somay Jain and Song Mei and Sonya Egoian and Spencer Kim and Spug Golden and SQ Mah and Steph Lin and Stephen Imm and Steve Sharpe and Steve Yadlowsky and Sulman Choudhry and Sungwon Eum and Suvansh Sanjeev and Tabarak Khan and Tal Stramer and Tao Wang and Tao Xin and Tarun Gogineni and Taya Christianson and Ted Sanders and Tejal Patwardhan and Thomas Degry and Thomas Shadwell and Tianfu Fu and Tianshi Gao and Timur Garipov and Tina Sriskandarajah and Toki Sherbakov and Tomek Korbak and Tomer Kaftan and Tomo Hiratsuka and Tongzhou Wang and Tony Song and Tony Zhao and Troy Peterson and Val Kharitonov and Victoria Chernova and Vineet Kosaraju and Vishal Kuo and Vitchyr Pong and Vivek Verma and Vlad Petrov and Wanning Jiang and Weixing Zhang and Wenda Zhou and Wenlei Xie and Wenting Zhan and Wes McCabe and Will DePue and Will Ellsworth and Wulfie Bain and Wyatt Thompson and Xiangning Chen and Xiangyu Qi and Xin Xiang and Xinwei Shi and Yann Dubois and Yaodong Yu and Yara Khakbaz and Yifan Wu and Yilei Qian and Yin Tat Lee and Yinbo Chen and Yizhen Zhang and Yizhong Xiong and Yonglong Tian and Young Cha and Yu Bai and Yu Yang and Yuan Yuan and Yuanzhi Li and Yufeng Zhang and Yuguang Yang and Yujia Jin and Yun Jiang and Yunyun Wang and Yushi Wang and Yutian Liu and Zach Stubenvoll and Zehao Dou and Zheng Wu and Zhigang Wang},
  year          = {2026},
  eprint        = {2601.03267},
  archiveprefix = {arXiv},
  primaryclass  = {cs.CL},
  url           = {https://arxiv.org/abs/2601.03267}
}

@misc{shao2024deepseekmathpushinglimitsmathematical,
  title         = {DeepSeekMath: Pushing the Limits of Mathematical Reasoning in Open Language Models},
  author        = {Zhihong Shao and Peiyi Wang and Qihao Zhu and Runxin Xu and Junxiao Song and Xiao Bi and Haowei Zhang and Mingchuan Zhang and Y. K. Li and Y. Wu and Daya Guo},
  year          = {2024},
  eprint        = {2402.03300},
  archiveprefix = {arXiv},
  primaryclass  = {cs.CL},
  url           = {https://arxiv.org/abs/2402.03300}
}

@inproceedings{rein2024gpqa,
  title     = {{GPQA}: A Graduate-Level Google-Proof {Q\&A} Benchmark},
  author    = {David Rein and Betty Li Hou and Asa Cooper Stickland and Jackson Petty and Richard Yuanzhe Pang and Julien Dirani and Julian Michael and Samuel R. Bowman},
  booktitle = {First Conference on Language Modeling},
  year      = {2024},
  url       = {https://openreview.net/forum?id=Ti67584b98}
}

@inproceedings{DBLP:conf/nips/WangMZNCGRAHJLK24,
  author={Yubo Wang and Xueguang Ma and Ge Zhang and Yuansheng Ni and Abhranil Chandra and Shiguang Guo and Weiming Ren and Aaran Arulraj and Xuan He and Ziyan Jiang and Tianle Li and Max Ku and Kai Wang and Alex Zhuang and Rongqi Fan and Xiang Yue and Wenhu Chen},
  title={MMLU-Pro: A More Robust and Challenging Multi-Task Language Understanding Benchmark},
  year={2024},
  cdate={1704067200000},
  url={http://papers.nips.cc/paper_files/paper/2024/hash/ad236edc564f3e3156e1b2feafb99a24-Abstract-Datasets_and_Benchmarks_Track.html},
  booktitle={NeurIPS}
}

@inproceedings{
brody2022how,
title={How Attentive are Graph Attention Networks? },
author={Shaked Brody and Uri Alon and Eran Yahav},
booktitle={International Conference on Learning Representations},
year={2022},
url={https://openreview.net/forum?id=F72ximsx7C1}
}

@inproceedings{10.5555/3454287.3455116,
  author    = {Ying, Rex and Bourgeois, Dylan and You, Jiaxuan and Zitnik, Marinka and Leskovec, Jure},
  title     = {GNNExplainer: generating explanations for graph neural networks},
  year      = {2019},
  publisher = {Curran Associates Inc.},
  address   = {Red Hook, NY, USA},
  booktitle = {Proceedings of the 33rd International Conference on Neural Information Processing Systems},
  articleno = {829},
  numpages  = {12}
}

@misc{grattafiori2024llama3herdmodels,
  title         = {The Llama 3 Herd of Models},
  author        = {Aaron Grattafiori and Abhimanyu Dubey and Abhinav Jauhri and Abhinav Pandey and Abhishek Kadian and Ahmad Al-Dahle and Aiesha Letman and Akhil Mathur and Alan Schelten and Alex Vaughan and Amy Yang and Angela Fan and Anirudh Goyal and Anthony Hartshorn and Aobo Yang and Archi Mitra and Archie Sravankumar and Artem Korenev and Arthur Hinsvark and Arun Rao and Aston Zhang and Aurelien Rodriguez and Austen Gregerson and Ava Spataru and Baptiste Roziere and Bethany Biron and Binh Tang and Bobbie Chern and Charlotte Caucheteux and Chaya Nayak and Chloe Bi and Chris Marra and Chris McConnell and Christian Keller and Christophe Touret and Chunyang Wu and Corinne Wong and Cristian Canton Ferrer and Cyrus Nikolaidis and Damien Allonsius and Daniel Song and Danielle Pintz and Danny Livshits and Danny Wyatt and David Esiobu and Dhruv Choudhary and Dhruv Mahajan and Diego Garcia-Olano and Diego Perino and Dieuwke Hupkes and Egor Lakomkin and Ehab AlBadawy and Elina Lobanova and Emily Dinan and Eric Michael Smith and Filip Radenovic and Francisco Guzmán and Frank Zhang and Gabriel Synnaeve and Gabrielle Lee and Georgia Lewis Anderson and Govind Thattai and Graeme Nail and Gregoire Mialon and Guan Pang and Guillem Cucurell and Hailey Nguyen and Hannah Korevaar and Hu Xu and Hugo Touvron and Iliyan Zarov and Imanol Arrieta Ibarra and Isabel Kloumann and Ishan Misra and Ivan Evtimov and Jack Zhang and Jade Copet and Jaewon Lee and Jan Geffert and Jana Vranes and Jason Park and Jay Mahadeokar and Jeet Shah and Jelmer van der Linde and Jennifer Billock and Jenny Hong and Jenya Lee and Jeremy Fu and Jianfeng Chi and Jianyu Huang and Jiawen Liu and Jie Wang and Jiecao Yu and Joanna Bitton and Joe Spisak and Jongsoo Park and Joseph Rocca and Joshua Johnstun and Joshua Saxe and Junteng Jia and Kalyan Vasuden Alwala and Karthik Prasad and Kartikeya Upasani and Kate Plawiak and Ke Li and Kenneth Heafield and Kevin Stone and Khalid El-Arini and Krithika Iyer and Kshitiz Malik and Kuenley Chiu and Kunal Bhalla and Kushal Lakhotia and Lauren Rantala-Yeary and Laurens van der Maaten and Lawrence Chen and Liang Tan and Liz Jenkins and Louis Martin and Lovish Madaan and Lubo Malo and Lukas Blecher and Lukas Landzaat and Luke de Oliveira and Madeline Muzzi and Mahesh Pasupuleti and Mannat Singh and Manohar Paluri and Marcin Kardas and Maria Tsimpoukelli and Mathew Oldham and Mathieu Rita and Maya Pavlova and Melanie Kambadur and Mike Lewis and Min Si and Mitesh Kumar Singh and Mona Hassan and Naman Goyal and Narjes Torabi and Nikolay Bashlykov and Nikolay Bogoychev and Niladri Chatterji and Ning Zhang and Olivier Duchenne and Onur Çelebi and Patrick Alrassy and Pengchuan Zhang and Pengwei Li and Petar Vasic and Peter Weng and Prajjwal Bhargava and Pratik Dubal and Praveen Krishnan and Punit Singh Koura and Puxin Xu and Qing He and Qingxiao Dong and Ragavan Srinivasan and Raj Ganapathy and Ramon Calderer and Ricardo Silveira Cabral and Robert Stojnic and Roberta Raileanu and Rohan Maheswari and Rohit Girdhar and Rohit Patel and Romain Sauvestre and Ronnie Polidoro and Roshan Sumbaly and Ross Taylor and Ruan Silva and Rui Hou and Rui Wang and Saghar Hosseini and Sahana Chennabasappa and Sanjay Singh and Sean Bell and Seohyun Sonia Kim and Sergey Edunov and Shaoliang Nie and Sharan Narang and Sharath Raparthy and Sheng Shen and Shengye Wan and Shruti Bhosale and Shun Zhang and Simon Vandenhende and Soumya Batra and Spencer Whitman and Sten Sootla and Stephane Collot and Suchin Gururangan and Sydney Borodinsky and Tamar Herman and Tara Fowler and Tarek Sheasha and Thomas Georgiou and Thomas Scialom and Tobias Speckbacher and Todor Mihaylov and Tong Xiao and Ujjwal Karn and Vedanuj Goswami and Vibhor Gupta and Vignesh Ramanathan and Viktor Kerkez and Vincent Gonguet and Virginie Do and Vish Vogeti and Vítor Albiero and Vladan Petrovic and Weiwei Chu and Wenhan Xiong and Wenyin Fu and Whitney Meers and Xavier Martinet and Xiaodong Wang and Xiaofang Wang and Xiaoqing Ellen Tan and Xide Xia and Xinfeng Xie and Xuchao Jia and Xuewei Wang and Yaelle Goldschlag and Yashesh Gaur and Yasmine Babaei and Yi Wen and Yiwen Song and Yuchen Zhang and Yue Li and Yuning Mao and Zacharie Delpierre Coudert and Zheng Yan and Zhengxing Chen and Zoe Papakipos and Aaditya Singh and Aayushi Srivastava and Abha Jain and Adam Kelsey and Adam Shajnfeld and Adithya Gangidi and Adolfo Victoria and Ahuva Goldstand and Ajay Menon and Ajay Sharma and Alex Boesenberg and Alexei Baevski and Allie Feinstein and Amanda Kallet and Amit Sangani and Amos Teo and Anam Yunus and Andrei Lupu and Andres Alvarado and Andrew Caples and Andrew Gu and Andrew Ho and Andrew Poulton and Andrew Ryan and Ankit Ramchandani and Annie Dong and Annie Franco and Anuj Goyal and Aparajita Saraf and Arkabandhu Chowdhury and Ashley Gabriel and Ashwin Bharambe and Assaf Eisenman and Azadeh Yazdan and Beau James and Ben Maurer and Benjamin Leonhardi and Bernie Huang and Beth Loyd and Beto De Paola and Bhargavi Paranjape and Bing Liu and Bo Wu and Boyu Ni and Braden Hancock and Bram Wasti and Brandon Spence and Brani Stojkovic and Brian Gamido and Britt Montalvo and Carl Parker and Carly Burton and Catalina Mejia and Ce Liu and Changhan Wang and Changkyu Kim and Chao Zhou and Chester Hu and Ching-Hsiang Chu and Chris Cai and Chris Tindal and Christoph Feichtenhofer and Cynthia Gao and Damon Civin and Dana Beaty and Daniel Kreymer and Daniel Li and David Adkins and David Xu and Davide Testuggine and Delia David and Devi Parikh and Diana Liskovich and Didem Foss and Dingkang Wang and Duc Le and Dustin Holland and Edward Dowling and Eissa Jamil and Elaine Montgomery and Eleonora Presani and Emily Hahn and Emily Wood and Eric-Tuan Le and Erik Brinkman and Esteban Arcaute and Evan Dunbar and Evan Smothers and Fei Sun and Felix Kreuk and Feng Tian and Filippos Kokkinos and Firat Ozgenel and Francesco Caggioni and Frank Kanayet and Frank Seide and Gabriela Medina Florez and Gabriella Schwarz and Gada Badeer and Georgia Swee and Gil Halpern and Grant Herman and Grigory Sizov and Guangyi and Zhang and Guna Lakshminarayanan and Hakan Inan and Hamid Shojanazeri and Han Zou and Hannah Wang and Hanwen Zha and Haroun Habeeb and Harrison Rudolph and Helen Suk and Henry Aspegren and Hunter Goldman and Hongyuan Zhan and Ibrahim Damlaj and Igor Molybog and Igor Tufanov and Ilias Leontiadis and Irina-Elena Veliche and Itai Gat and Jake Weissman and James Geboski and James Kohli and Janice Lam and Japhet Asher and Jean-Baptiste Gaya and Jeff Marcus and Jeff Tang and Jennifer Chan and Jenny Zhen and Jeremy Reizenstein and Jeremy Teboul and Jessica Zhong and Jian Jin and Jingyi Yang and Joe Cummings and Jon Carvill and Jon Shepard and Jonathan McPhie and Jonathan Torres and Josh Ginsburg and Junjie Wang and Kai Wu and Kam Hou U and Karan Saxena and Kartikay Khandelwal and Katayoun Zand and Kathy Matosich and Kaushik Veeraraghavan and Kelly Michelena and Keqian Li and Kiran Jagadeesh and Kun Huang and Kunal Chawla and Kyle Huang and Lailin Chen and Lakshya Garg and Lavender A and Leandro Silva and Lee Bell and Lei Zhang and Liangpeng Guo and Licheng Yu and Liron Moshkovich and Luca Wehrstedt and Madian Khabsa and Manav Avalani and Manish Bhatt and Martynas Mankus and Matan Hasson and Matthew Lennie and Matthias Reso and Maxim Groshev and Maxim Naumov and Maya Lathi and Meghan Keneally and Miao Liu and Michael L. Seltzer and Michal Valko and Michelle Restrepo and Mihir Patel and Mik Vyatskov and Mikayel Samvelyan and Mike Clark and Mike Macey and Mike Wang and Miquel Jubert Hermoso and Mo Metanat and Mohammad Rastegari and Munish Bansal and Nandhini Santhanam and Natascha Parks and Natasha White and Navyata Bawa and Nayan Singhal and Nick Egebo and Nicolas Usunier and Nikhil Mehta and Nikolay Pavlovich Laptev and Ning Dong and Norman Cheng and Oleg Chernoguz and Olivia Hart and Omkar Salpekar and Ozlem Kalinli and Parkin Kent and Parth Parekh and Paul Saab and Pavan Balaji and Pedro Rittner and Philip Bontrager and Pierre Roux and Piotr Dollar and Polina Zvyagina and Prashant Ratanchandani and Pritish Yuvraj and Qian Liang and Rachad Alao and Rachel Rodriguez and Rafi Ayub and Raghotham Murthy and Raghu Nayani and Rahul Mitra and Rangaprabhu Parthasarathy and Raymond Li and Rebekkah Hogan and Robin Battey and Rocky Wang and Russ Howes and Ruty Rinott and Sachin Mehta and Sachin Siby and Sai Jayesh Bondu and Samyak Datta and Sara Chugh and Sara Hunt and Sargun Dhillon and Sasha Sidorov and Satadru Pan and Saurabh Mahajan and Saurabh Verma and Seiji Yamamoto and Sharadh Ramaswamy and Shaun Lindsay and Shaun Lindsay and Sheng Feng and Shenghao Lin and Shengxin Cindy Zha and Shishir Patil and Shiva Shankar and Shuqiang Zhang and Shuqiang Zhang and Sinong Wang and Sneha Agarwal and Soji Sajuyigbe and Soumith Chintala and Stephanie Max and Stephen Chen and Steve Kehoe and Steve Satterfield and Sudarshan Govindaprasad and Sumit Gupta and Summer Deng and Sungmin Cho and Sunny Virk and Suraj Subramanian and Sy Choudhury and Sydney Goldman and Tal Remez and Tamar Glaser and Tamara Best and Thilo Koehler and Thomas Robinson and Tianhe Li and Tianjun Zhang and Tim Matthews and Timothy Chou and Tzook Shaked and Varun Vontimitta and Victoria Ajayi and Victoria Montanez and Vijai Mohan and Vinay Satish Kumar and Vishal Mangla and Vlad Ionescu and Vlad Poenaru and Vlad Tiberiu Mihailescu and Vladimir Ivanov and Wei Li and Wenchen Wang and Wenwen Jiang and Wes Bouaziz and Will Constable and Xiaocheng Tang and Xiaojian Wu and Xiaolan Wang and Xilun Wu and Xinbo Gao and Yaniv Kleinman and Yanjun Chen and Ye Hu and Ye Jia and Ye Qi and Yenda Li and Yilin Zhang and Ying Zhang and Yossi Adi and Youngjin Nam and Yu and Wang and Yu Zhao and Yuchen Hao and Yundi Qian and Yunlu Li and Yuzi He and Zach Rait and Zachary DeVito and Zef Rosnbrick and Zhaoduo Wen and Zhenyu Yang and Zhiwei Zhao and Zhiyu Ma},
  year          = {2024},
  eprint        = {2407.21783},
  archiveprefix = {arXiv},
  primaryclass  = {cs.AI},
  url           = {https://arxiv.org/abs/2407.21783}
}

@misc{qwen2025qwen25technicalreport,
  title         = {Qwen2.5 Technical Report},
  author        = {Qwen and An Yang and Baosong Yang and Beichen Zhang and Binyuan Hui and Bo Zheng and Bowen Yu and Chengyuan Li and Dayiheng Liu and Fei Huang and Haoran Wei and Huan Lin and Jian Yang and Jianhong Tu and Jianwei Zhang and Jianxin Yang and Jiaxi Yang and Jingren Zhou and Junyang Lin and Kai Dang and Keming Lu and Keqin Bao and Kexin Yang and Le Yu and Mei Li and Mingfeng Xue and Pei Zhang and Qin Zhu and Rui Men and Runji Lin and Tianhao Li and Tianyi Tang and Tingyu Xia and Xingzhang Ren and Xuancheng Ren and Yang Fan and Yang Su and Yichang Zhang and Yu Wan and Yuqiong Liu and Zeyu Cui and Zhenru Zhang and Zihan Qiu},
  year          = {2025},
  eprint        = {2412.15115},
  archiveprefix = {arXiv},
  primaryclass  = {cs.CL},
  url           = {https://arxiv.org/abs/2412.15115}
}

@misc{xiaomi2025mimo,
  title         = {{MiMo}: Unlocking the Reasoning Potential of Language Model -- From
                   Pretraining to Posttraining},
  author        = {{Xiaomi LLM-Core Team}},
  year          = {2025},
  eprint        = {2505.07608},
  archiveprefix = {arXiv},
  primaryclass  = {cs.CL},
  url           = {https://arxiv.org/abs/2505.07608}
}

@inproceedings{wang2024mathshepherd,
  title     = {Math-Shepherd: Verify and Reinforce {LLMs} Step-by-step without Human
               Annotations},
  author    = {Wang, Peiyi and Li, Lei and Shao, Zhihong and Xu, Runxin and Dai, Damai
               and Li, Yifei and Chen, Deli and Wu, Yu and Sui, Zhifang},
  booktitle = {Proceedings of the 62nd Annual Meeting of the Association for
               Computational Linguistics (Volume 1: Long Papers)},
  pages     = {9426--9439},
  year      = {2024}
}

@inproceedings{hu2019strategies,
  title     = {Strategies for Pre-training Graph Neural Networks},
  author    = {Hu, Weihua and Liu, Bowen and Gomes, Joseph and Zitnik, Marinka and
               Liang, Percy and Pande, Vijay S and Leskovec, Jure},
  booktitle = {International Conference on Learning Representations},
  year      = {2020},
  url       = {https://openreview.net/forum?id=HJlWWJSFDH}
}

@inproceedings{
lightman2024lets,
title={Let's Verify Step by Step},
author={Hunter Lightman and Vineet Kosaraju and Yuri Burda and Harrison Edwards and Bowen Baker and Teddy Lee and Jan Leike and John Schulman and Ilya Sutskever and Karl Cobbe},
booktitle={The Twelfth International Conference on Learning Representations},
year={2024},
url={https://openreview.net/forum?id=v8L0pN6EOi}
}

@misc{liu2026skyworkrewardv2scalingpreferencedata,
      title={Skywork-Reward-V2: Scaling Preference Data Curation via Human-AI Synergy},
      author={Chris Yuhao Liu and Liang Zeng and Yuzhen Xiao and Jujie He and Jiacai Liu and Chaojie Wang and Rui Yan and Wei Shen and Fuxiang Zhang and Jiacheng Xu and Yang Liu and Yahui Zhou},
      year={2026},
      eprint={2507.01352},
      archivePrefix={arXiv},
      primaryClass={cs.CL},
      url={https://arxiv.org/abs/2507.01352},
}

@inproceedings{
xu2018how,
title={How Powerful are Graph Neural Networks?},
author={Keyulu Xu and Weihua Hu and Jure Leskovec and Stefanie Jegelka},
booktitle={International Conference on Learning Representations},
year={2019},
url={https://openreview.net/forum?id=ryGs6iA5Km},
}

@inproceedings{neves-ribeiro-etal-2022-entailment,
    title = "Entailment Tree Explanations via Iterative Retrieval-Generation Reasoner",
    author = "Neves Ribeiro, Danilo  and
      Wang, Shen  and
      Ma, Xiaofei  and
      Dong, Rui  and
      Wei, Xiaokai  and
      Zhu, Henghui  and
      Chen, Xinchi  and
      Xu, Peng  and
      Huang, Zhiheng  and
      Arnold, Andrew  and
      Roth, Dan",
    editor = "Carpuat, Marine  and
      de Marneffe, Marie-Catherine  and
      Meza Ruiz, Ivan Vladimir",
    booktitle = "Findings of the Association for Computational Linguistics: NAACL 2022",
    month = jul,
    year = "2022",
    address = "Seattle, United States",
    publisher = "Association for Computational Linguistics",
    url = "https://aclanthology.org/2022.findings-naacl.35/",
    doi = "10.18653/v1/2022.findings-naacl.35",
    pages = "465--475"
}

@inproceedings{10448355,
  author    = {Mo, Shentong and Xin, Miao},
  booktitle = {ICASSP 2024 -- 2024 IEEE International Conference on Acoustics, Speech and Signal Processing ({ICASSP})},
  title     = {Tree of Uncertain Thoughts Reasoning for Large Language Models},
  year      = {2024},
  pages     = {12742--12746},
  doi       = {10.1109/ICASSP48485.2024.10448355}
}

@inproceedings{linjiaen-etal-2025-optimizing,
    title = "Optimizing Multi-Hop Document Retrieval Through Intermediate Representations",
    author = "Lin, Jiaen  and
      Liu, Jingyu  and
      Liu, Yingbo",
    editor = "Che, Wanxiang  and
      Nabende, Joyce  and
      Shutova, Ekaterina  and
      Pilehvar, Mohammad Taher",
    booktitle = "Findings of the Association for Computational Linguistics: ACL 2025",
    month = jul,
    year = "2025",
    address = "Vienna, Austria",
    publisher = "Association for Computational Linguistics",
    url = "https://aclanthology.org/2025.findings-acl.816/",
    doi = "10.18653/v1/2025.findings-acl.816",
    pages = "15798--15809",
    ISBN = "979-8-89176-256-5"
}

@misc{weng2026learning,
  title        = {Learning beyond gradients},
  author       = {Weng, Jiayi},
  year         = {2026},
  url          = {https://trinkle23897.github.io/learning-beyond-gradients/},
  note         = {Blog post, \url{https://trinkle23897.github.io/learning-beyond-gradients/}, accessed September 2026}
}

@misc{fey2019fastgraphrepresentationlearning,
  title={Fast Graph Representation Learning with PyTorch Geometric},
  author={Matthias Fey and Jan Eric Lenssen},
  year={2019},
  eprint={1903.02428},
  archivePrefix={arXiv},
  primaryClass={cs.LG},
  url={https://arxiv.org/abs/1903.02428}
}

\appendix
\newpage
\tableofcontents
\newpage
\section{CoT2Graph Implementation Details}
\label{app:implementation}

This appendix provides implementation details of the CoT2Graph framework, covering the full process from LLM hyperparameters to the graph construction pipeline and GNN configuration.

\subsection{LLM Generation Hyperparameters}
\label{app:gen_hyperparams}

All experiments share the same generation configuration across all models (DS-Qw7B, DS-Qw14B, MiMo-7B-RL) and datasets (GPQA, MMLU-Pro) to ensure comparability. Inference was performed locally using the \texttt{vLLM} engine and weights were loaded in \texttt{bfloat16} precision. Experiments were run on NVIDIA A100, RTX 3090 and 4090 GPUs.
We use a low-temperature decoding setup with temperature $T = 0.1$, a maximum generation length of $2^{14} \approx 16K$ tokens, batch size of 16, and fixed random seed 42. These hyperparameters are chosen to favour stable outputs while preserving stochasticity for diversity in Best-of-$N$ evaluation.
During generation, we also collect the top-$K = 5$ token log-probabilities to compute uncertainty metrics.

\subsection{Node Segmentation}
Raw CoT text is segmented into atomic steps using a keyword-based parser.
Transition markers (\textit{Step}, \textit{Wait}, \textit{So}, \textit{Finally},
\textit{Let me verify}, \textit{However}, \textit{Therefore}, and related phrases) are
identified via case-insensitive regex matching similar to previous work~\cite{jiang-etal-2025-makes}.
Segments shorter than 100 characters are merged with the following step to ensure semantic
completeness.
The resulting sequence $v_1, v_2, \ldots, v_n$ preserves linear generation order.
Parser robustness and the embedding-based alternative are described in Appx.~\ref{app:segmentation_review}.

\subsection{Node Feature Vector}
Let $n=|\mathcal{V}|$. Each node $v_i$ is represented by a six-dimensional feature vector~\citep{jiang-etal-2025-makes}:
\begin{multline*}
\phi(v_i) = \bigl[i,\; i/n,\; d^{\mathrm{in}}_i/n,\; d^{\mathrm{out}}_i/n,\\
        d^{\mathrm{in}}_i - d^{\mathrm{out}}_i,\; d^{\mathrm{in}}_i/(d^{\mathrm{out}}_i +
        1)\bigr],
\end{multline*}
encoding absolute and normalized position, in-/out-degree, and degree balance and ratio.

\subsection{Edge Classification}
Dependency relationships are inferred by Llama-3.1-8B-Instruct~\cite{grattafiori2024llama3herdmodels}
applied to the upper triangular half of the adjacency matrix: for all pairs $(v_i, v_j)$
with $i < j$, the model returns an edge type $t \in \mathcal{T}$ and a strength $s \in [0, 10]$.
Restricting to the upper triangle reduces prompt cost and encodes a causal prior consistent with
sequential generation of the LLM. Appx.~\ref{app:prompts} includes the prompt used for edge classification.

The six edge types are:
\textit{inference} (strict logical or mathematical deduction from $v_i$ to $v_j$),
\textit{background} (prior knowledge from $v_i$ contextualizing $v_j$),
\textit{observation} (direct empirical support from $v_i$ for $v_j$),
\textit{meta-reflection} (reasoning about $v_i$'s validity that informs $v_j$),
\textit{repetition} ($v_i$ and $v_j$ express the same content or a semantic rephrasing), and
\textit{none} (no meaningful dependency).

\subsection{Graph Compression via Union-Find}
Nodes connected by \textit{repetition} edges are merged with Union--Find. Each connected component becomes a canonical supernode, and its incident edges are redirected through the node-to-representative map. This removes repeated formulations while preserving their dependencies.

\subsection{GNN Architecture and Training}
Implementation uses PyTorch Geometric~\cite{fey2019fastgraphrepresentationlearning}.
\paragraph{Graph-Level Outcome Prediction.}
The GATv2~\cite{brody2022how} classifier comprises two GATv2Conv
layers (hidden dimension 64, single attention head, edge features incorporated via attention
weighting), global mean pooling, and a two-layer MLP head
(Linear($64\!\to\!64$)--ReLU--Dropout(0.2)--Linear($64\!\to\!2$)).
Training uses Adam ($\mathrm{lr} = 10^{-3}$, weight decay $10^{-4}$), batch size 32,
StepLR scheduling (step 20, $\gamma = 0.5$), early stopping (patience 50), and 200 maximum
epochs.

\paragraph{Node-Level Classifiers.}
The GINE~\cite{hu2019strategies} classifier uses a two-layer encoder (hidden dimension 128)
with multi-task prediction heads, cosine annealing scheduling, and 300 maximum epochs.
GINE's sum aggregation is preferred over attention-based pooling for node-level tasks because
it provides stronger inductive bias for distinguishing local neighborhood configurations.

\paragraph{Training Protocol.}
Graph-level results use 11 trace-level splits with seed 11. For the 545 GPQA graphs, each split assigns 445 graphs to training and validation and 100 to testing; 15\% of the former are held out for validation before training-set class balancing. Early stopping selects validation F1. Node-level experiments use 400 training and 145 test graphs over 11 splits, and Best-of-$N$ a question-level 70/30 split with seed 42, reserving 20\% of training questions for validation.

\section{\texorpdfstring{CoT Segmentation Analysis}{CoT Segmentation Analysis}}
\label{app:segmentation_review}

\subsection{\texorpdfstring{Rule-Based Segmentation Sensitivity}{Rule-Based Segmentation Sensitivity}}

We use a regex parser to avoid an additional GPT-5 Mini call for every trace. For each model, its transition markers were obtained by combining human inspection with GPT-5 Mini keyword extraction on a subset of traces. Their consistency across questions and datasets may reflect shared post-training conventions, so the marker list is model-specific.

We parsed all 545 GPQA traces from DS-Qw7B, obtaining 62.1 nodes per trace on average, and judged them against the GPT-5 Mini reference segmentation. Across these traces, 91.5\% of parser nodes matched, 8.4\% were undersplit, and fewer than 0.1\% were oversplit. At trace level, 79.3\% had at most 5\% incorrectly segmented nodes and 86.7\% had at most 10\%. The 100-character merge favors preserving context over fragmenting a coherent step: for example, markers in a short span of the form ``FIRST ... SO ...'' can denote one argument rather than two steps.

Coding tasks were not evaluated. On MATH500 the parser stays reliable because reasoning models pair every derivation with prose carrying the same markers. The DS-Qw7B step below opens with \textit{Now} (in bold) and already exceeds the merge threshold, so the parser keeps it as one node.

\begin{Verbatim}[frame=single, framesep=2.5mm, fontsize=\small,
  commandchars=\\\{\},
  label={\normalfont\small\textbf{MATH500\textsubscript{ID 404} --- DS-Qw7B}}]
\textbf{Now}, determine the number of boxes
with neither pens nor pencils.
Subtract the number of boxes with pens
or pencils from the total:
Neither = Total boxes - |P or Q|
Neither = 10 - 7 = 3
\end{Verbatim}

Thresholds and markers may still require dataset- and model-specific tuning.

\subsection{\texorpdfstring{Embedding-Based Segmentation}{Embedding-Based Segmentation}}

We considered semantic segmentation before adopting the regex parser. CoT segmentation is underdetermined: adjacent sentences may be separate derivation steps, connective tissue, or fragments of one calculation. In pilot runs, embedding neighborhoods often tracked sentence length and recurring discourse markers more than reasoning content. We tested K-means on sentence embeddings, dynamic programming with a within-segment similarity objective and complexity penalty, multi-scale boundary scoring with Ward-style merging, and an optimal-coverage objective over contiguous windows. We also tried gradient-free policy search in the sense of~\citet{weng2026learning}. None of these variants produced boundaries stable enough for the graph analyses, so all reported results use the rule-based parser.

\section{\texorpdfstring{Edge Type Definitions and Validation}{Edge Type Definitions and Validation}}
\label{app:edge_type_definitions}

\subsection{\texorpdfstring{Definitions and Literature Support}{Definitions and Literature Support}}
\label{app:edge_type_literature}

The edge annotator assigns one of five relations, or \textit{none}, to each ordered node pair A and B:
\begin{itemize}[leftmargin=*]
    \item \textbf{Inference}: Node A is an explicit premise from which Node B follows directly.
    \item \textbf{Background information}: Node A contains prior knowledge, context, assumptions, or general facts that are used in Node B.
    \item \textbf{Observation}: Node B is an observation, comment, or general step related to Node A, even indirectly.
    \item \textbf{Meta-reflection}: Node B plans, checks, or reflects on the reasoning process rather than advancing the solution.
    \item \textbf{Repetition}: Nodes A and B express the same or a closely similar idea, result, or reasoning step.
    \item \textbf{None}: A and B have no meaningful relation.
\end{itemize}

The annotator also assigns a strength score: 0 for no link; 1--3 for a weak relation; 4--6 for a moderate but nonessential contribution; 7--8 when Node A is a primary premise; and 9--10 when Node B follows necessarily from Node A alone.

The taxonomy combines inference and support relations from entailment trees~\cite{dalvi-etal-2021-explaining,neves-ribeiro-etal-2022-entailment}, background knowledge from retrieval-based reasoning~\cite{linjiaen-etal-2025-optimizing}, and reflection and repetition from search-based reasoning~\cite{10.5555/3666122.3666639,10448355,jiang-etal-2025-makes}. Graph of Thoughts motivates representing these relations as edges between thought nodes~\cite{10.1609/aaai.v38i16.29720}.

\subsection{\texorpdfstring{Semantic Equivalence and Conflict Resolution}{Semantic Equivalence and Conflict Resolution}}
\label{app:edge_merge_review}
Two nodes are semantically equivalent when they express the same intermediate conclusion or a rephrasing with the same meaning. The annotator marks this relation as \textit{repetition}; Union--Find merges each connected repetition component into a supernode and redirects its incident edges. Overall, 10.8\% of nodes participate in such components.

Redirected edges can assign different types to the same pair of supernodes. We retain the edge with the highest LLM strength; equal-strength ties use inference $>$ meta-reflection $>$ observation $>$ background. This order follows the mean annotated strength by type across the evaluated models and datasets and agrees with the GNNExplainer attribution ranking. Among tied-strength comparisons, 75.1\% are trivial because both edges already have the same type, since they link to nodes with equivalent content.

\subsection{\texorpdfstring{Human Annotation Validation}{Human Annotation Validation}}
\label{app:annotation_validation_review}
We validate edge labels on three GPQA CoTs generated by DS-Qw14B, containing 60, 61, and 49 nodes (170 total). The evaluated edges are sampled with stratification across judge-assigned edge classes: few calibration examples shown with the judge's label and 95 blind examples. Three annotators label each edge independently: one paper author, one external participant familiar with LLM evaluation, and Fable Claude. Their majority vote gives one ensemble label per edge.

The judge is one annotation under comparison, not a gold standard. Since sampling is stratified by its labels, the row totals of Table~\ref{tab:judge_vs_ensemble} are fixed by design and do not show how often each relation occurs. Table~\ref{tab:pairwise_agreement} shows that judge--ensemble agreement is 85.3\% with Cohen's $\kappa=0.82$, while Table~\ref{tab:judge_vs_ensemble} compares the two class by class.

\vspace{0.3em}
\begin{Verbatim}[frame=single, framesep=3mm, fontsize=\footnotesize,
  label={\normalfont\small\textbf{Human Reviewer Instructions}}]
TASK: Follow and read the definitions below
(see: Prompt 3 - Edge Classification). For
each ordered node pair, select one relation
type, assign confidence on a 1--5 scale,
and optionally add a comment.
\end{Verbatim}

\begin{table}[H]
    \centering
    \scriptsize
    \setlength{\tabcolsep}{4pt}
    \renewcommand{\arraystretch}{1.15}
    \begin{tabular}{lccccc}
        \toprule
         & \textbf{Judge} & \textbf{Author} & \textbf{Extern.} & \textbf{Fable} & \textbf{Ens.} \\
        \midrule
        Judge (Llama-3.1-8B) & -- & 86.3\% & 81.1\% & 84.2\% & 85.3\% \\
        Author               & 0.83 & -- & 84.2\% & 87.4\% & 94.7\% \\
        External             & 0.77 & 0.81 & -- & 82.1\% & 89.5\% \\
        Fable Claude         & 0.80 & 0.84 & 0.78 & -- & 92.6\% \\
        Ensemble             & 0.82 & 0.93 & 0.87 & 0.91 & -- \\
        \bottomrule
    \end{tabular}
    \caption{Pairwise agreement between annotations: raw agreement above the diagonal, Cohen's $\kappa$ below.}
    \label{tab:pairwise_agreement}
\end{table}

\begin{table}[!t]
    \centering
    \scriptsize
    \setlength{\tabcolsep}{2.5pt}
    \renewcommand{\arraystretch}{1.15}
    \begin{tabular}{lcccccc@{\hskip 7pt}c}
        \toprule
        \textbf{Judge $\backslash$ Ens.} & \textit{Inf} & \textit{Bkg} & \textit{Obs} & \textit{Meta} & \textit{Rep} & \textit{None} & \textbf{Recall} \\
        \midrule
        Inference        & \textbf{18} &   1 &   1 &   0 &   0 &   0 & 90.0\% \\
        Bkg. info        &   0 & \textbf{16} &   0 &   1 &   0 &   2 & 84.2\% \\
        Observation      &   0 &   1 & \textbf{17} &   2 &   0 &   1 & 81.0\% \\
        Meta-refl.       &   0 &   0 &   0 & \textbf{13} &   1 &   1 & 86.7\% \\
        Repetition       &   0 &   0 &   0 &   0 & \textbf{1} &   0 & 100.0\% \\
        None             &   0 &   1 &   1 &   1 &   0 & \textbf{16} & 84.2\% \\
        \midrule
        $n$                &  18 &  19 &  19 &  17 &   2 &  20 & 95 \\
        \textbf{Precision} & 100.0\% & 84.2\% & 89.5\% & 76.5\% & 50.0\% & 80.0\% & \\
        \textbf{F1}        & 0.95 & 0.84 & 0.85 & 0.81 & 0.67 & 0.82 & \\
        \bottomrule
    \end{tabular}
    \caption{LLM-as-Judge against the annotator ensemble on the sampled edges. Cells are edge counts; recall and precision are the diagonal over the row and column totals. Accuracy 85.3\%, Cohen's $\kappa=0.82$. \BLUE{\textit{Repetition} is near-absent here because it concentrates in long traces (\S\ref{subsec:efficiency}), which are out of scope and too costly to annotate.}}
    \label{tab:judge_vs_ensemble}
\end{table}

\section{\texorpdfstring{Ablation Studies}{Ablation Studies}}
\label{app:ablation_review}
Table~\ref{tab:gnn_ablations} separates model baselines from edge, node-feature, and topology ablations on the same fixed splits; each ablation removes one candidate signal source. Logistic regression and random forest use graph size, density, and degree summaries; the pooled MLP removes message passing; and the sequence encoder removes graph edges. The untyped GNN preserves connectivity and strength while replacing the edge-type vector with a constant.

The graph-statistics logistic regression performs below the length-only baseline, indicating that coarse summaries alone do not explain the full model's gain. All variants are run on GPQA traces from DS-Qw7B over the same 11 fixed splits; reported values are accuracy means with standard deviations across splits. Edge types are masked one channel at a time: edges stay in the graph and reach the GNN with a neutral type, so connectivity is unchanged and only type signal is lost. Their contributions are additive only if the remaining channels are unaffected.

Removing the \textit{inference} channel produces the largest single-type drop ($63.0$), followed by \textit{background} ($64.2$) and \textit{observation} ($65.8$), whereas masking \textit{meta-reflection} changes the score least ($67.6$). Removing the degree features costs more than removing position ($66.4$ against $68.3$).

\begin{table*}[!t]
    \centering
    \scriptsize
    \setlength{\tabcolsep}{3.2pt}
    \renewcommand{\arraystretch}{1.2}
    \begin{tabular}{l l p{5.4cm} c c}
        \toprule
        \textbf{Group} & \textbf{Method / variant} & \textbf{Controlled input change}
        & \textbf{Accuracy (\%)} & $\Delta$ \textbf{vs. ours} \\
        \midrule
        \multirow{6}{*}{Baseline}
          & Graph-statistics RF & same summaries as LR & $58.3$ \scriptsize{$\pm4.0$} & $-12.2$ \\
          & Graph-statistics LR & size, density, degree summaries & $60.5$ \scriptsize{$\pm4.5$} & $-10.0$ \\
          & \underline{Length-only LR} & CoT length only & $\underline{65.9}$ \scriptsize{$\pm3.1$} & $-4.6$ \\
          & Sequence encoder & ordered node features, no graph edges & $66.1$ \scriptsize{$\pm3.6$} & $-4.4$ \\
          & \underline{LCoT2Tree}~\cite{jiang-etal-2025-makes} & LLM-annotated tree sketch instead of the dependency graph & $\underline{68.2}$ \scriptsize{$\pm2.8$} & $-2.3$ \\
          & Pooled-feature MLP & pooled node/edge features, no message passing & $68.4$ \scriptsize{$\pm4.1$} & $-2.1$ \\
        \midrule
        \multirow{7}{*}{Edge ablation}
          & Without inference type & mask inference channel only & $63.0$ \scriptsize{$\pm3.4$} & $-7.5$ \\
          & Without background type & mask background channel only & $64.2$ \scriptsize{$\pm4.4$} & $-6.3$ \\
          & Without observation type & mask observation channel only & $65.8$ \scriptsize{$\pm4.6$} & $-4.7$ \\
          & Untyped-edge GNN & keep connectivity/strength; replace type one-hot by a constant & $66.1$ \scriptsize{$\pm4.8$} & $-4.4$ \\
          & Without repetition type & mask repetition channel only & $67.5$ \scriptsize{$\pm3.8$} & $-3.0$ \\
          & Without meta-reflection type & mask meta-reflection channel only & $67.6$ \scriptsize{$\pm5.1$} & $-2.9$ \\
          & Without edge strength & keep topology and edge types & $68.6$ \scriptsize{$\pm5.0$} & $-1.9$ \\
        \midrule
        \multirow{2}{*}{Node ablation}
          & Without degree features & remove in/out degree, balance, and ratio & $66.4$ \scriptsize{$\pm4.3$} & $-4.1$ \\
          & Without position & remove absolute/normalized position & $68.3$ \scriptsize{$\pm5.3$} & $-2.2$ \\
        \midrule
        Extension & Semantic node embeddings & add frozen sentence embeddings to structural features & $64.8$ \scriptsize{$\pm5.5$} & $-5.7$ \\
        \midrule
        \textbf{Reference} & \textbf{CoT2Graph} (GATv2 GNN) & graph topology + edge type + strength + node features & $\mathbf{70.5}$ \scriptsize{$\pm2.3$} & $0.0$ \\
        \bottomrule
    \end{tabular}
    \caption{Controlled baselines and ablations for final-answer prediction on GPQA (DS-Qw7B), accuracy (\%) over the same 11 fixed splits, mean $\pm$ s.d. Rows are sorted by increasing score within each group; $\Delta$ is against CoT2Graph. The \textbf{bold} row is our CoT2Graph; the two \underline{underlined} rows are the length LR and LCoT2Tree baselines of Table~\ref{tab:gnn_results}.
    Abbrev.: LR (logistic regression), RF (random forest), MLP (multi-layer perceptron), GNN (graph neural network).}
    \label{tab:gnn_ablations}
\end{table*}

\section{\texorpdfstring{Computational Scaling Analysis}{Computational Scaling Analysis}}
\label{app:cost_review}
CoT2Graph is suitable for offline analysis of fixed traces, while online use is limited by pairwise edge annotation.
Edge annotation dominates the cost of the pipeline. For a trace with $n$ nodes it queries Llama-3.1-8B-Instruct on at most $n(n-1)/2$ ordered pairs. Pairs are batched and traces are distributed across GPUs, without assuming ideal linear speedup. On 545 GPQA DS-Qw7B traces with a mean of 62 nodes, annotation takes 2.28 minutes per trace on an RTX 3090; the corresponding A100 and H100 figures, 0.21 and 0.12 minutes, are extrapolations rather than measurements. A log--log regression of annotation time on trace length gives an exponent of 1.31, below the $O(n^2)$ worst case.

The remaining stages are negligible next to annotation. Graph compression and feature extraction are linear in the number of edges, and the GATv2 classifier has fewer than 15K parameters and trains on 545 graphs, so it takes negligible GPU time.

\section{Cross-Model Generalization}
\label{app:cross_model}

\begin{table*}[!t]
	\centering
	\scriptsize
	\setlength{\tabcolsep}{4pt}
	\renewcommand{\arraystretch}{1.2}
	\begin{tabular}{l r l l p{5.2cm}}
    	\toprule
    	\textbf{Model} & \textbf{Size} & \textbf{Base} & \textbf{Post-training} & \textbf{Key Characteristics} \\
    	\midrule
    	DS-Qw7B
    	& 7B & Qwen2.5-Math-7B & SFT on distilled CoTs (no direct RL)
    	& $\sim$800k traces from DeepSeek-R1 teacher \\
    	DS-Qw14B
    	& 14B & Qwen2.5-14B & SFT on distilled CoTs (no direct RL)
    	& Same pipeline; larger capacity for long-context \\
    	MiMo-7B-RL
    	& 7B & MiMo (Xiaomi) & Direct RL with verifiable rewards
    	& 130k math/code problems; difficulty-aware sampling \\
    	\bottomrule
    \end{tabular}
        \caption{%
    		Summary of the three reasoning models evaluated in the cross-model generalization study.
    		Models are selected to provide controlled variation in scale and post-training strategy.%
    	}
    	\label{tab:model_summary}
    \end{table*}
    
\definecolor{darkgreen}{RGB}{0,120,0}
\definecolor{darkred}{RGB}{150,0,0}
\begin{table*}[!t]
	\centering
	\scriptsize
	\setlength{\tabcolsep}{4pt}
	\renewcommand{\arraystretch}{1.2}
	\begin{tabular}{llccc}
		\toprule
		\textbf{Method (F1 score)}
		 & \textbf{Train ($\downarrow$) / Test ($\to$)}
		 & \textbf{DS-Qw7B}
		 & \textbf{DS-Qw14B}
		 & \textbf{MiMo-7B-RL}                                                             \\
		\midrule
		\multirow{3}{*}{LCoT2Tree~\cite{jiang-etal-2025-makes}}
		 & DS-Qw7B     & {59.4} & {56.5} & 46.6          \\
		 & DS-Qw14B    & 57.8          & 56.1          & 42.6          \\
		 & MiMo-7B-RL                      & 56.6          & 52.3          & 42.9          \\
		\midrule
		\multirow{3}{*}{CoT2Graph \textit{(ours)}}
         & DS-Qw7B  
         & {62.0 (+2.6)} 
         & {56.6 (+0.1)} 
         & {50.0 (+3.4)} \\
        
         & DS-Qw14B 
         & {57.3 (-0.5)} 
         & {58.2 (+2.1)} 
         & {45.8 (+3.2)} \\
        
         & MiMo-7B-RL 
         & {60.7 (+4.1)} 
         & {54.1 (+1.8)} 
         & {51.5 (+8.6)} \\
		\bottomrule
	\end{tabular}
    \caption{%
        Cross-model generalization: macro F1 (\%) for graph-level correctness prediction on GPQA.
        Rows indicate the LLM that generated the CoT traces used to train the GNN predictor model, while columns indicate the CoT traces used for testing.
        CoT2Graph improves over LCoT2Tree in most settings, with gains up to +8.6\% F1.
        }
	\label{tab:cross_model_f1}
\end{table*}

We ask whether topology learned from one generator transfers to another.
A structurally grounded CoT representation should, in principle, capture reasoning patterns
that transcend the idiosyncrasies of any particular model; the graph topology induced
by correct vs.\ incorrect reasoning should share universal features regardless of which model
produced the trace.
We test this hypothesis through a cross-model transfer study, training the
GNN classifier on traces from one model and evaluating on traces from another.
Beyond validating our method, this analysis reveals how post-training
and model scale shape the underlying reasoning structure.

\subsection{Model Characterization}
\label{app:model_char}

We evaluate three models chosen to span controlled variation across two key axes:
\emph{model scale} (7B vs.\ 14B) and \emph{post-training strategy} (knowledge distillation
vs.\ direct reinforcement learning).
Table~\ref{tab:model_summary} summarizes their principal characteristics.

\paragraph{DS-Qw7B \& 14B.}
Both models are initialized from the Qwen2.5 base models~\cite{qwen2025qwen25technicalreport} and trained exclusively via
supervised fine-tuning (SFT) on approximately 800k reasoning traces distilled from the
DeepSeek-R1 teacher model~\cite{Guo_2025}, itself trained with Group Relative Policy Optimization
(GRPO)~\cite{shao2024deepseekmathpushinglimitsmathematical} through a four-stage pipeline (cold-start SFT $\to$ RL $\to$ SFT $\to$ RL
alignment).
Critically, the DS-Qw7B and DS-Qw14B distillation models receive \emph{no direct RL signal}; their
reasoning behavior is shaped entirely by behavioral cloning from the teacher's outputs.

\paragraph{MiMo-7B-RL.}
MiMo~\cite{xiaomi2025mimo} is quite distant from DeepSeek.
Its pre-training curriculum spans three stages with
multi-token prediction objectives, making reasoning performance a first-class training objective.
Post-training applies direct reinforcement learning on 130k verifiable mathematics and
programming problems, using difficulty-aware reward sampling.

\subsection{Cross-Model Transfer Results}
\label{app:transfer}

Table~\ref{tab:cross_model_f1} reports F1 scores for graph-level correctness classification.
We report macro F1 as the comparison metric in this analysis rather than accuracy because class imbalance varies across
models, reflecting their differing baseline accuracy on GPQA.

\paragraph{Overall Generalization.}
CoT2Graph consistently outperforms LCoT2Tree in cross-model transfer.
The most pronounced improvements occur on MiMo-7B-RL traces: within-model ($51.5\%$ vs.\
$42.9\%$, $+8.6\%$) and in DS-Qw7B $\to$ MiMo-7B-RL transfer ($50.0\%$ vs.\ $46.6\%$,
$+3.4\%$).
These results confirm that graph-based structural representations encode reasoning
patterns that generalize more broadly across models than features derived from
LLM-generated reasoning tree sketches.
Moreover, LCoT2Tree requires a frontier model (e.g.,
DeepSeek-v3) to produce the intermediate sketch used for tree construction, introducing a
costly and model-dependent preprocessing step.
CoT2Graph, by contrast, uses only a fixed 7B edge classifier, a substantially more
scalable and deployment-friendly architecture.

\paragraph{Same Model Family, Different Scale. 
}
Transfer within the DeepSeek distillation family is moderate and approximately symmetric:
$57.3\%$ (train DS-Qw14B, test DS-Qw7B) vs.\ $57.8\%$ (train DS-Qw7B, test DS-Qw14B) for CoT2Graph.
Both models are trained with the same distillation pipeline from the same teacher, so their
reasoning graphs share the same high-level structural vocabulary.
The slight asymmetry (training on DS-Qw7B generalizes marginally better to DS-Qw14B
than vice versa) may reflect that 7B traces, being somewhat simpler in structure,
produce features that transfer across the capacity gap more robustly.

\paragraph{Different Model Family, Matched Scale. 
}
The most informative comparison is between the two 7B models, which share scale but
diverge in post-training strategy.
Transfer is notably asymmetric: training on MiMo-7B-RL and testing on DS-Qw7B yields $60.7\%$,
while the reverse yields only $50.0\%$.
This asymmetry reveals that RL-trained models produce structurally richer reasoning
graphs whose patterns subsume those of distillation-trained models.
A GNN trained on MiMo's RL-driven traces (with greater topological diversity,
more frequent exploration loops, and more varied synthesis node configurations) learns a
generalizable structural vocabulary of correct and incorrect reasoning.
Conversely, one trained on the more uniform distillation traces of DS-Qw7B transfers
less well to MiMo-7B-RL's broader structural patterns.

\paragraph{The Role of Post-Training Strategy.}
MiMo-7B-RL's lower within-model F1 ($51.5\%$) compared to DS-Qw7B ($62.0\%$) and DS-Qw14B
($58.2\%$) reflects the \emph{higher structural diversity of RL-trained traces}: greater
topological variance makes binary classification harder in absolute terms, but provides
richer training signal for cross-model transfer.

\section{Optimal TSP* Path Details}
\label{app:TSP}
\begin{figure}[H]
	\centering
    \text{Suboptimal Discovery Inefficiency}
    \includegraphics[width=0.92\linewidth, trim=0 0 0 1.7em, clip]{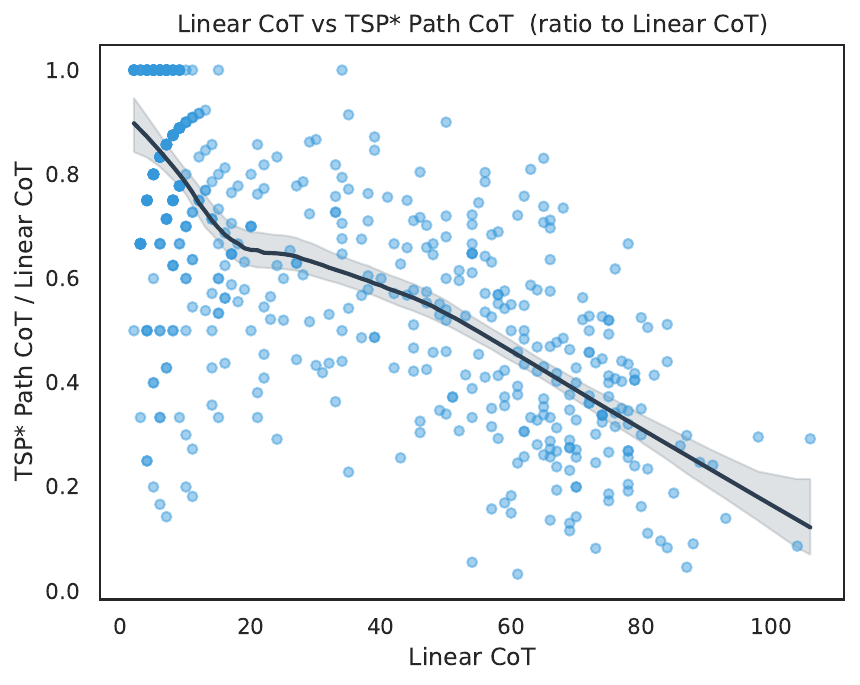}
	\caption{%
    	Scatter plot of loop-free CoT length ($L_{\mathrm{loops}}$, x-axis) vs.\ inefficiency-based efficiency
    	($E = 1 - \mathcal{L}_{\mathrm{ineff.}}$, y-axis), where $\mathcal{L}_{\mathrm{ineff.}}$ captures only path inefficiency
    	relative to $L_{\mathrm{TSP^*}}$ and excludes $L_{\mathrm{loops}}$ redundancy.
	}
	\label{fig:efficiency_TSP}
\end{figure}

We detail the path construction used to estimate $L_{\mathrm{TSP^*}}$, covering the traversal and the relaxation applied when a circular dependency blocks progress.
Algorithm~\ref{alg:TSP} estimates $L_{\mathrm{TSP^*}}$ with a dependency-consistent traversal. A blocked traversal removes the edge with the lowest LLM strength. Strength is the primary criterion; only equal-strength ties use the removal order background $>$ observation $>$ meta-reflection $>$ inference. The importance order follows the mean annotated strength by type across the evaluated models and datasets and is consistent with the GNNExplainer attribution ranking. Appx.~\ref{app:edge_merge_review} reports the related tie analysis, and Figure~\ref{fig:efficiency_TSP} relates loop-free CoT length to $\mathcal{L}_{\mathrm{ineff.}}$, the path inefficiency measured against $L_{\mathrm{TSP^*}}$.

{\removelatexerror
\begin{algorithm}[H]
	\small
	\DontPrintSemicolon
	\SetAlgoNlRelativeSize{-1}
	\KwIn{CoT graph $\mathcal{G} = (\mathcal{V}, \mathcal{E})$;
		LLM-as-Judge relevance labels per node}
	\KwOut{Ordered path $\mathcal{P}$; path length $L_{\mathrm{TSP^*}}$; relaxation count $\tau$}
	Remove all nodes labelled \textit{redundant} or \textit{irrelevant}\;
	$\mathcal{S} \leftarrow \{v_0\}$ \tcp*{visited; $v_0$ = problem node}
	$\mathcal{R} \leftarrow \mathcal{V} \setminus \{v_0\}$;\quad
	$\mathcal{P} \leftarrow [v_0]$;\quad $\tau \leftarrow 0$\;
	\While{$\mathcal{R} \neq \emptyset$}{
		$\mathcal{F} \leftarrow
			\bigl\{v \in \mathcal{R} \;\big|\;
			\forall (u,v) \in \mathcal{E}: u \in \mathcal{S}\bigr\}$
		\tcp*{prerequisites satisfied}
		\eIf{$\mathcal{F} \neq \emptyset$}{
			$v^* \leftarrow \arg\min_{v \in \mathcal{F}} \operatorname{id}(v)$\;
			Append $v^*$ to $\mathcal{P}$;\quad
			$\mathcal{S} \leftarrow \mathcal{S} \cup \{v^*\}$;\quad
			$\mathcal{R} \leftarrow \mathcal{R} \setminus \{v^*\}$\;
		}{
			\tcp{deadlock: relax prereq.}
			Remove the minimum-strength blocking edge $(u,v)$ with $v \in \mathcal{R}$ and $u \notin \mathcal{S}$\;
			\hspace{1em}Break ties by removing \textit{background}, \textit{observation},
			\textit{meta-reflection}, then \textit{inference}\;
			$\tau \leftarrow \tau + 1$\;
		}
	}
	\Return $\mathcal{P}$, $L_{\mathrm{TSP^*}} = |\mathcal{P}|$, $\tau$\;
	\caption{Graph-Based Optimal Path Construction (TSP-Inspired)}
	\label{alg:TSP}
\end{algorithm}}

\section{\texorpdfstring{Additional Figures and Tables}{Additional Figures and Tables}}
\label{app:additional_results}
\label{app:node_hallucination_metrics}
\begin{figure}[H]
	\centering
	\includegraphics[width=0.92\linewidth]{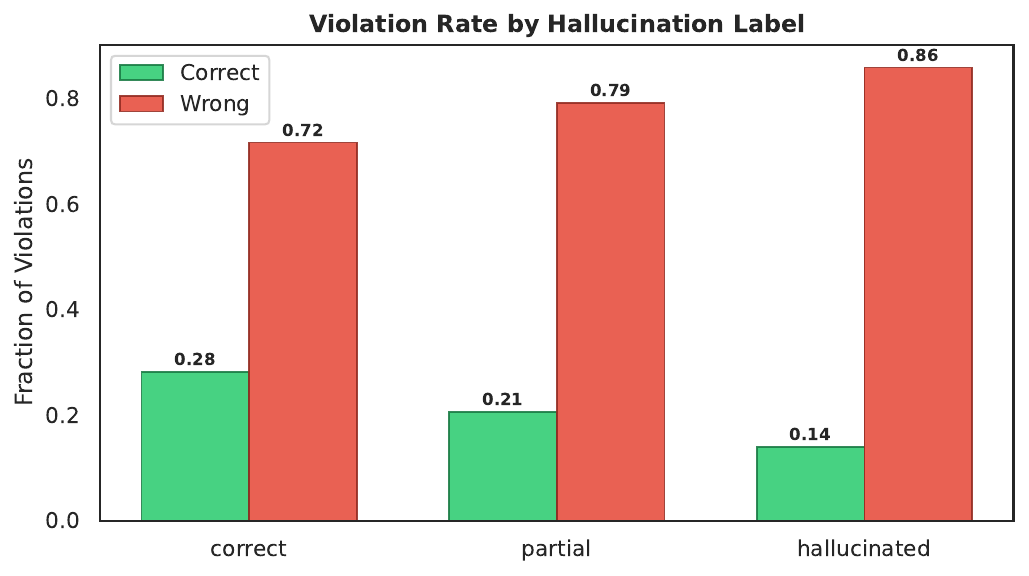}
	\caption{%
        Comparison of step-skipping violation rate (see \S\ref{subsec:efficiency}) across node categories (\textit{correct, partial, hallucinated}) using LLM-as-Judge annotations (GPQA, DS-Qw14B).
        For each category, the bar is split between correct (green) and incorrect CoTs (red); similar results are obtained with Qwen-7B.
    }
	\label{fig:hallucination_vs_stepskipping}
\end{figure}

\begin{figure}[H]
	\centering
	\includegraphics[width=\linewidth]{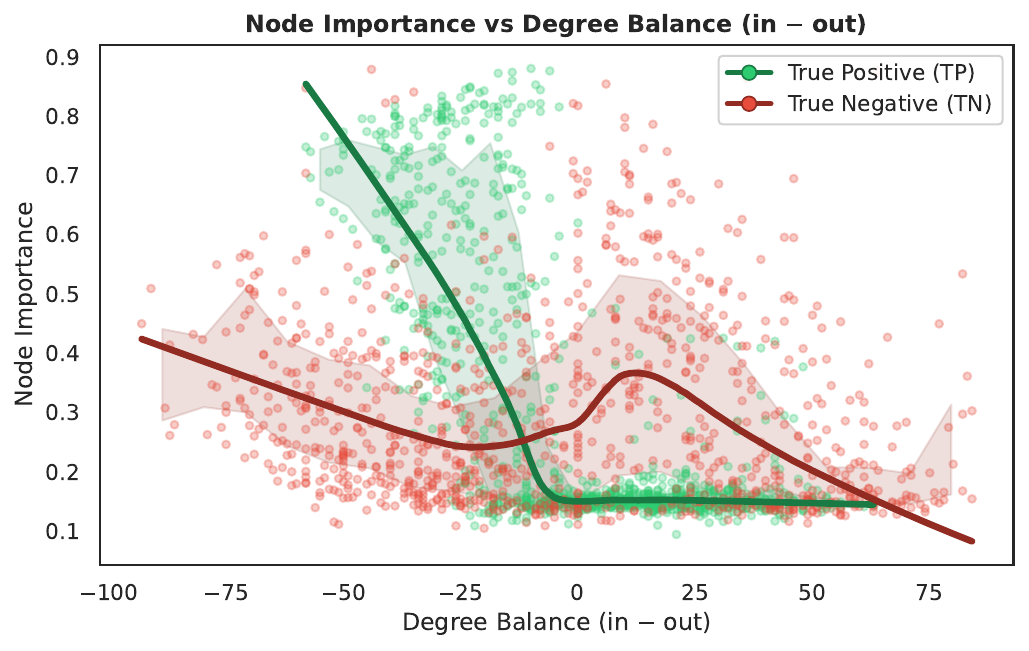}
	\caption{%
    	GNNExplainer node importance vs.\ (in-out)-degree values (GPQA, DS-Qw7B).
    }
	\label{fig:node_importance_degree}
\end{figure}

\begin{figure}[H]
	\centering
    \text{Best-of-$N$ Decoding Correlation}
	\includegraphics[width=\linewidth]{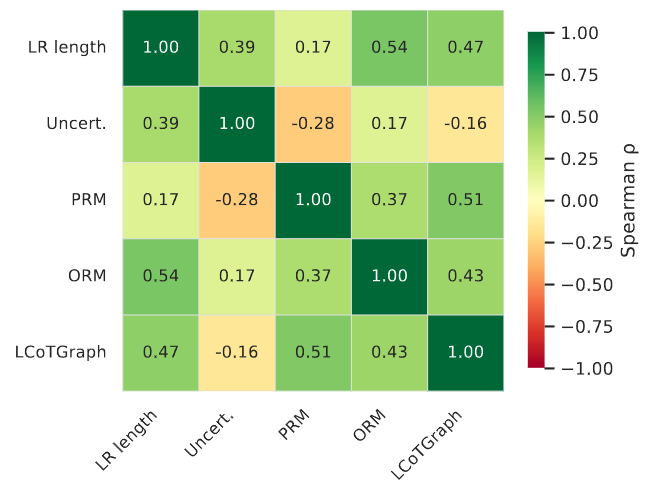}
	\caption{%
		Spearman rank correlation matrix of Best-of-$N$ selection scores on MMLU-Pro
		(DS-Qw7B, $N{=}5$). 
	}
	\label{fig:bestofN_corr}
\end{figure}

\begin{figure*}[!t]
	\centering
    \textbf{CoT2Graph: CoT $\to$ Reasoning Structure}
	\includegraphics[width=\linewidth]{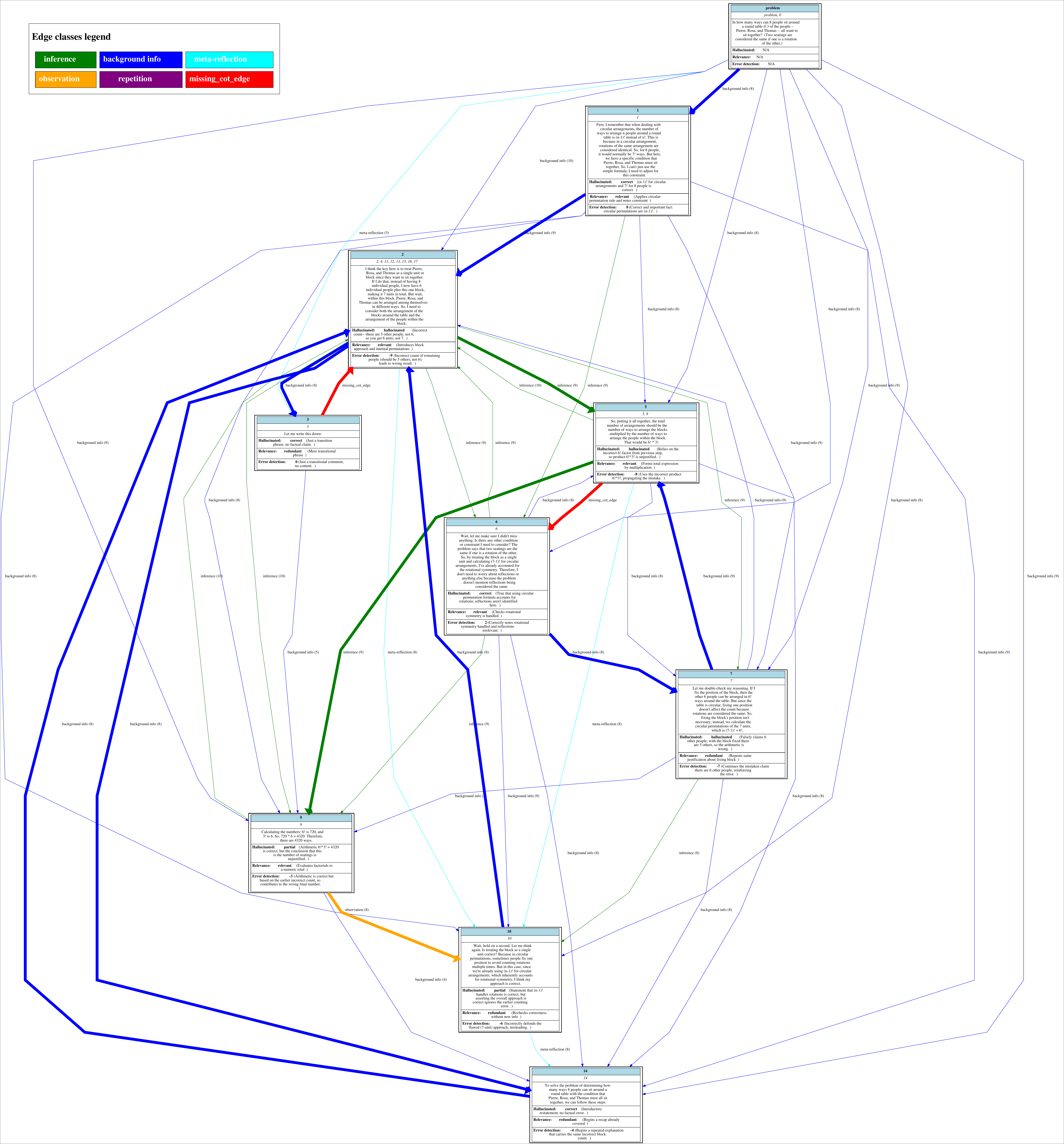}
	\caption{%
		Full dependency graph for a representative GPQA (dataset entry ID: 32) trace
        (DS-Qw14B, correct prediction).
        Nodes correspond to atomic reasoning steps; edge colours encode type
        (\textit{inference}: green, \textit{background-info}: blue, \textit{observation}: yellow,
        \textit{meta-reflection}: light blue, \textit{missing-edge}: red).
        Repetition edges are not shown because their endpoints are merged into representative supernodes. Node 5, for example, represents the merged set $\{5,8\}$ and marks a return to previously expressed content.
	}
	\label{fig:cot_graph}
\end{figure*}

This appendix maps the additional diagnostics: Figure~\ref{fig:hallucination_vs_stepskipping} covers step skipping, Table~\ref{tab:relevance_cm} covers relevance, Figure~\ref{fig:node_importance_degree} covers attribution, Figure~\ref{fig:bestofN_corr} covers Best-of-$N$ complementarity, and Figure~\ref{fig:cot_graph} shows the full example graph.

The node-level hallucination accuracies reported in \S\ref{subsec:node_analysis} are aggregated over the 11 fixed splits described in Appx.~\ref{app:implementation}, using the same GINE model as the relevance task.

\begin{table}[H]
	\centering
	\small
	\setlength{\tabcolsep}{6pt}
	\renewcommand{\arraystretch}{1.15}
	\begin{tabular}{lccc}
		\toprule
		\textbf{True $\backslash$ Pred}
		           & \textit{Relevant}
		           & \textit{Redundant}
		           & \textit{Irrelevant}                       \\
		\midrule
		Relevant   & \textbf{65.2}       & 34.8          & 0.0 \\
		Redundant  & 33.5                & \textbf{66.5} & 0.0 \\
		Irrelevant & 21.1                & 78.9          & \textbf{0.0} \\
		\bottomrule
	\end{tabular}
    \caption{%
		Confusion matrix (\%) for node-level relevance prediction on GPQA
		(DS-Qw14B).
		Rows are ground-truth; columns are predicted labels.%
	}
	\label{tab:relevance_cm}
\end{table}

\section{Prompts}
\label{app:prompts}
The following prompts are used for CoT generation, LLM-as-Judge evaluation, and graph edge classification: Prompt~1 generates the CoT traces, Prompts~2a--c define the judge targets, and Prompt~3 defines graph edges.

\vspace{0.3em}
\begin{Verbatim}[frame=single, framesep=3mm, fontsize=\footnotesize,
  label={\normalfont\small\textbf{Prompt 1 --- CoT Generation}}]
Solve the problem step by step.

Write the full reasoning process, with
each step separated clearly from the
next. Show all intermediate reasoning
needed to reach the solution.

After completing the reasoning, write
the final answer clearly at the end
and do not generate additional text.

Problem:
{question}

Options:
A. {option_1}
B. {option_2}
...
\end{Verbatim}

\vspace{0.3em}
\begin{Verbatim}[frame=single, framesep=3mm, fontsize=\footnotesize,
  label={\normalfont\small\textbf{Prompt 2a --- LLM-as-Judge: Hallucination}}]
TASK: Check each numbered step for
factual accuracy, independently of
other steps.

LABELS:
- correct: node fully accurate
- partial: node partially accurate,
  minor errors or incomplete
- hallucinated: factually wrong or
  unsupported

INPUT:
Problem: {problem}
Expected Solution: {solution}
Expected Final Answer: {answer}
Reasoning Steps: {cot}

INSTRUCTIONS: Evaluate all steps
0 to {cot_len-1} independently.

OUTPUT (ONE LINE PER NODE):
[i] <correct|partial|hallucinated>
    | WHY: <3 words>
\end{Verbatim}

\vspace{0.3em}
\begin{Verbatim}[frame=single, framesep=3mm, fontsize=\footnotesize,
  label={\normalfont\small\textbf{Prompt 2b --- LLM-as-Judge: Relevance}}]
TASK: Assess each step's contribution
to the derivation of the final answer.

LABELS:
- relevant: contributes new information
- redundant: repeats prior information
- irrelevant: unrelated to solution

INPUT:
Problem: {problem}
Expected Solution: {solution}
Expected Final Answer: {answer}
Reasoning Steps: {cot}

INSTRUCTIONS: Analyze all steps
0 to {cot_len-1} in order.

OUTPUT (ONE LINE PER NODE):
[i] <relevant|redundant|irrelevant>
    | WHY: <3 words>
\end{Verbatim}

\begin{Verbatim}[frame=single, framesep=3mm, fontsize=\footnotesize,
  label={\normalfont\small\textbf{Prompt 2c --- LLM-as-Judge: Final Answer}}]
TASK: Determine if the trace reaches
the correct answer.

LABELS:
- final_answer_correct:
  <Correct, Incorrect>
- final_answer_in_cot: step numbers
  containing the final answer

INPUT:
Problem: {problem}
Expected Solution: {solution}
Expected Final Answer: {answer}
Reasoning Steps: {cot}

OUTPUT FORMAT (STRICT):
Final answer correct:
  <Correct, Incorrect>
Final answer in CoT:
  [comma-separated step numbers]
\end{Verbatim}

\begin{Verbatim}[frame=single, framesep=3mm, fontsize=\footnotesize,
  label={\normalfont\small\textbf{Prompt 3 - Edge Classification}}]
You are given the original problem statement
and two nodes from a reasoning process that
attempts to solve it:

Problem Statement: {problem_text}

Node A (#{src_idx}): {src_text}
Node B (#{tgt_idx}): {tgt_text}

Your task: Evaluate the relationship from
Node A to Node B. If there is a doubt about
any relationship, choose "none" with
strength 0.

Decide how Node A contributes to Node B by
choosing exactly one of the following
relationship types:
- "repetition": Node A and Node B express
  substantially the same or a very similar
  idea, result, conclusion, or reasoning
  step, even if phrased differently.
- "background info": Node A contains prior
  knowledge, context, assumptions, or
  general facts that are used in Node B.
- "inference": Node A provides an explicit
  premise, and Node B is just the direct
  logical consequence of it. Consider
  background info, observation, and
  meta-reflection instead if more
  appropriate.
- "observation": Node B is an observation,
  comment, or general step related to
  Node A, even indirectly.
- "meta-reflection": Node B involves task
  understanding, planning, decomposition,
  verification, self-checking, or reasoning
  about the problem-solving process rather
  than advancing the solution itself.
- "none": No meaningful relationship
  between Node A and Node B. This should be
  the default choice whenever the
  relationship is unclear, weak,
  speculative, or indirect.

Assign a strength score from 0 to 10:
- 0: no link - unrelated nodes
- 1-3: weak - indirect relation; Node A
  barely influences Node B
- 4-6: moderate - Node A contributes to
  Node B but is not required for it
- 7-8: strong - Node A is a primary driver
  or direct premise of Node B
- 9-10: essential - Node B directly and
  necessarily follows from Node A alone

Output exactly one valid JSON object, with
no extra text:
{"class": <"meta-reflection" |
 "repetition" | "inference" |
 "background info" | "observation" |
 "none">, "strength": <0-10>}

So, the JSON that classifies the pair is
the following:
\end{Verbatim}



\definecolor{light-blue}{RGB}{0,191,255}

\clearpage
\end{document}